\documentclass[10pt,twocolumn,letterpaper]{article}

\usepackage{3dv}

\makeatletter
\@namedef{ver@everyshi.sty}{}
\makeatother

\usepackage{times}
\usepackage{epsfig}
\usepackage{graphicx}
\usepackage{grffile}
\usepackage{amsmath}
\usepackage{amssymb}

\usepackage{graphicx}
\usepackage{comment}
\usepackage{subcaption}

\usepackage{placeins}

\usepackage{epsfig}
\usepackage{tabularx}
\usepackage{helvet}
\usepackage{adjustbox}
\usepackage{booktabs}
\usepackage{bm}

\makeatletter

\usepackage{tikz}
\usetikzlibrary{arrows,shapes,positioning}

\usepackage[pagebackref=true,breaklinks=true,letterpaper=true,colorlinks,bookmarks=false]{hyperref}
\usepackage{cleveref}

\newcommand{\High}{\mathrm{H}}
\newcommand{\Low}{\mathrm{L}}

\newcommand{\Color}{\bm{C}}
\newcommand{\Depth}{\bm{D}}
\newcommand{\T}{T_{\High \rightarrow \Low}}
\newcommand{\D}{\Delta_{\High \rightarrow \Low}}
\newcommand{\Ts}{t}

\newcommand{\twodots}{\mathinner {\ldotp \ldotp}}
\newcommand{\intrange}[2]{[#1\twodots #2]}

\newcommand{\U}{M^1}
\renewcommand{\L}{M^2}

\newcommand{\mutlirow}[2]{\begin{tabular}{@{}c@{}}#1 \\ #2\end{tabular}}

\threedvfinalcopy 

\def\threedvPaperID{283} 

\ifthreedvfinal\fi

\newcommand\blfootnote[1]{%
  \begingroup
  \renewcommand\thefootnote{}\footnote{#1}%
  \addtocounter{footnote}{-1}%
  \endgroup
}

\begin{document}

\title{ Self-supervised Depth Denoising Using Lower- and Higher-quality RGB-D sensors}


\author{Akhmedkhan Shabanov$^1$ \and Ilya Krotov$^1$
\and Nikolay Chinaev$^1$ \and Vsevolod Poletaev$^1$
\and Sergei Kozlukov$^2$ \and Igor Pasechnik$^3$
\and Bulat Yakupov$^1$ \and Artsiom Sanakoyeu$^4$
\and Vadim Lebedev$^1$ \and Dmitry Ulyanov$^1$}



\maketitle


\begin{abstract}
Consumer-level depth cameras and depth sensors embedded in mobile devices enable numerous applications, such as AR games and face identification. However, the quality of the captured depth is sometimes insufficient for 3D reconstruction, tracking and other computer vision tasks. 
In this paper, we propose a self-supervised depth denoising approach to denoise and refine depth coming from a low quality sensor.
We record simultaneous RGB-D sequences with unzynchronized lower- and higher-quality cameras and solve a challenging problem of aligning sequences both temporally and spatially. We then learn a deep neural network to denoise the lower-quality depth using the matched higher-quality data as a source of supervision signal. 
We experimentally validate our method against state-of-the-art filtering-based and deep denoising techniques and show its application for 3D object reconstruction tasks where our approach leads to more detailed fused surfaces and better tracking.
\end{abstract}

\section{Introduction}


\begin{figure}
    \centering
    \begin{tabular}{@{}cccc@{}}
    \includegraphics[width=.20\textwidth]{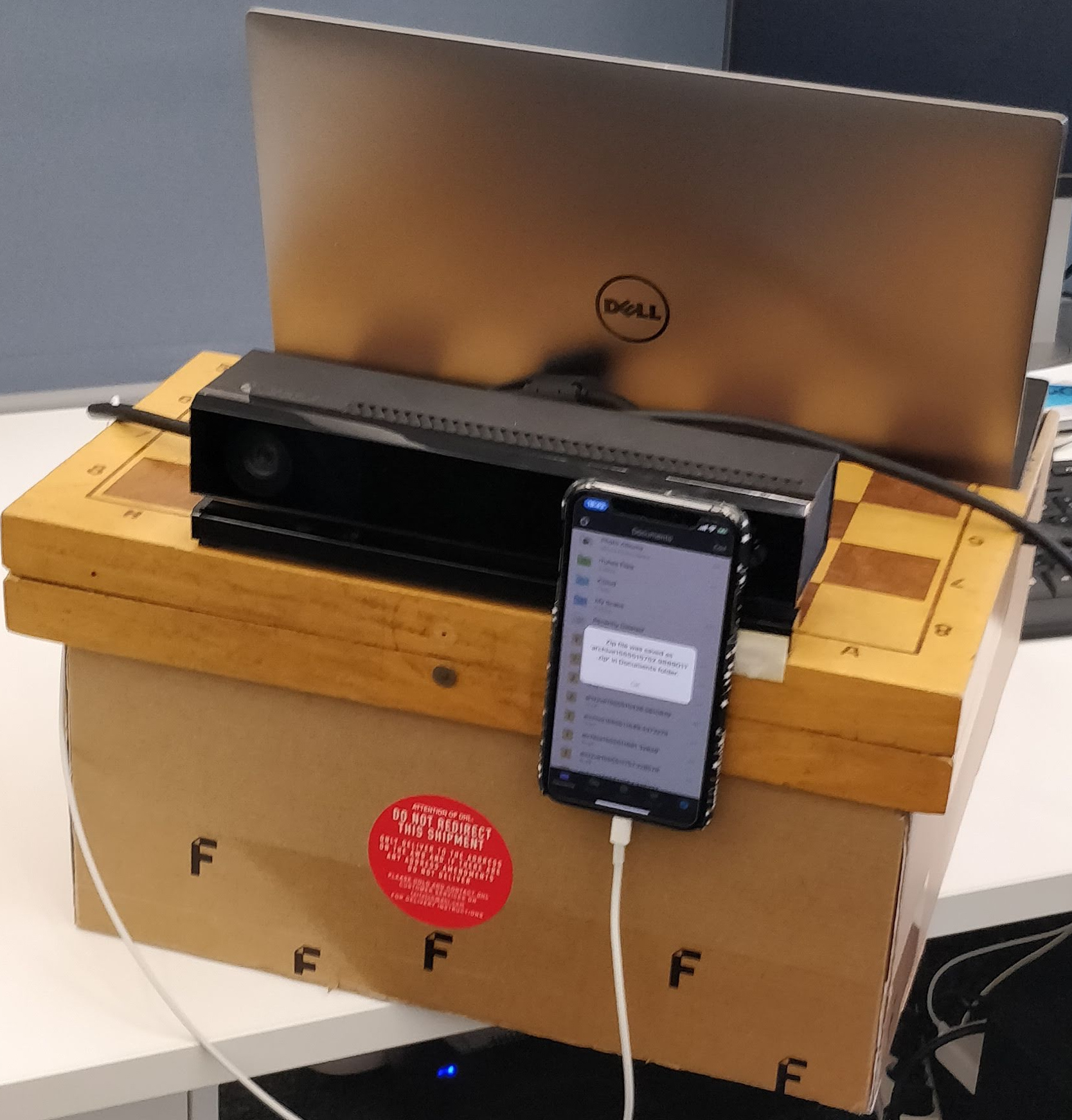} &
    \includegraphics[width=.10\textwidth]{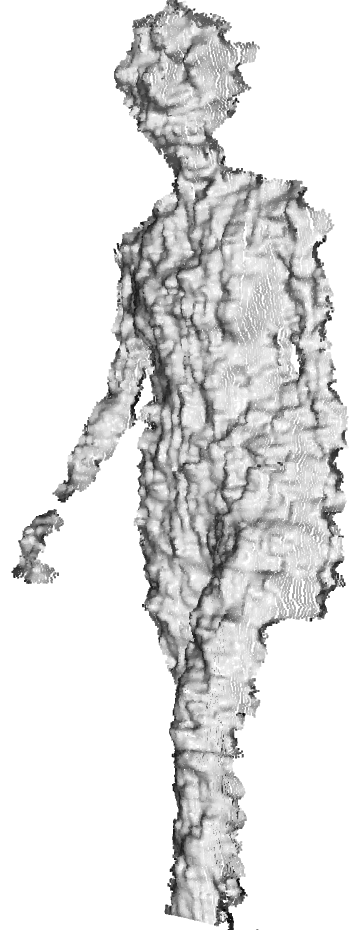} &
    \includegraphics[width=.10\textwidth]{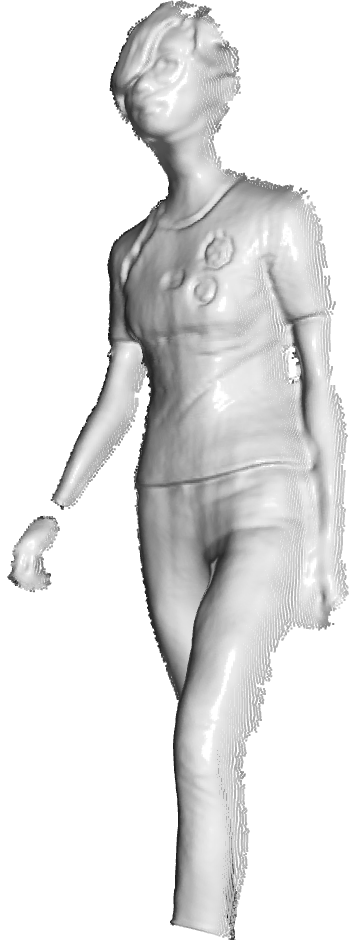} \\
    (a) & (b) LQ depth & (c) Ours
    \end{tabular}
    \vspace*{3mm}\caption{We use lower- and higher-quality depth sensors (a) to record simultaneous RGB-D sequences. Our automatic data collection pipeline allows us to train a model for denoising the data coming from the lower-quality (LQ) sensor (b). In figure (c) we show a result produced by our method for the input (b).}\label{fig:teaser}
\end{figure}

\blfootnote{$^1$In3D, $^2$Skoltech, $^3$Samsung AI Center, Moscow, $^4$Heidelberg Collaboratory for Image Processing.\\
Contact e-mail: shabanov.ae@phystech.edu}
A depth sensor is an extremely valuable tool for many computer vision applications, and the latest generations of smartphones often include one. These sensors provide an important cue about the three-dimensional world, supplementing the output of a standard RGB camera. Depth data is used to create a bokeh effect, to perform 3D reconstruction, to enable a more immersive experience in  augmented/virtual reality and games. Unfortunately, the full utilization of depth sensors is sometimes complicated by their insufficient accuracy. \Cref{fig:teaser} (b) shows an example of a depth map captured using a TrueDepth camera of an iPhone X. 

Low signal-to-noise ratio of mobile depth sensors prevents the use of the existing well-developed computer vision methods~\cite{newcombe2011kinectfusion,newcombe2015dynamicfusion,Dai2017bundlefusion,yu2018doublefusion} that are designed to work with more accurate sensors, such as Microsoft Kinect V2~\cite{sell2014xbox}. Therefore, there is a need to intelligently denoise the depth coming from mobile sensors.
Classic denoising methods based on filtering~\cite{Tomasi1998bf,Kopf07jbf,Zhang2014rgf} are not able to work well with noise of such magnitude and lead to over-smoothing and loss of details. The development of data-driven and supervised approaches is complicated as it is highly non-trivial and sometimes even impossible to collect ground truth data for supervision. Recently, \cite{Yan2018ddr,Jeon2018,HyeokHyenKwon2015} proposed to synthesize ground truth by adopting depth fusion methods~\cite{newcombe2011kinectfusion,Dai2017bundlefusion,yu2018doublefusion}. To overcome the lack of ground truth data, Sterzentsenko \etal~\cite{sterzentsenko2019self} used multiple sensors and self-supervised photometric reprojection loss to train a deep neural network.

In this paper we propose a learning-based method for depth denoising of lower-quality (LQ) depth sensor's output using supervision of a higher-quality (HQ) depth sensor.
We use a very simple but flexible method to acquire ground truth data. We place LQ and HQ sensors nearby and record a set of simultaneous RGB-D sequences. We consider an in-the-wild scenario where the hardware clock synchronization and prior extrinsic calibration for the sensors is not possible. We perform automatic alignment of the recorded sequences both in time and space by estimating the time delta and an extrinsic transform between the devices.
We then use the frames shot around the same moment in time to train a deep network to transform a depth image from the LQ sensor to a corresponding depth image from the HQ sensor reprojected to the LQ camera view. Note that our self-supervised approach extracts supervision signal automatically from the data and does not require any human labeling. 

While the proposed method can be used with any pair of RGB-D sensors, in this work, we aim to denoise lower-quality Apple TrueDepth (TD) camera and use Microsoft Kinect V2 (K2) as supervision. We focus on an application of 3D body scanning challenging to do with a TD sensor. In fact, designed with the task of face identification in mind, the TD camera is tuned for close objects, and quality of the produced data quickly deteriorates at the distance of 1-2 meters that is required to fit a standing person into the field of view. Therefore, we create a dataset with people captured in a variety of poses and lightning conditions using a rig shown in~\cref{fig:teaser}. To our knowledge, our dataset is the first to include RGB-D sequences recorded simultaneously with different quality RGB-D sensors.


We extensively test the proposed method against state-of-the-art filtering and deep-learning based algorithms. We demonstrate the effectiveness of our learned denoising in 3D surface reconstruction applications. Developed for sensors such as K2, KinectFusion~\cite{newcombe2011kinectfusion} and DoubleFusion~\cite{yu2018doublefusion} fail to work with noisy sensors such as TD and our approach enables usage of these reconstruction algorithms without any modifications. 

To summarize, our contributions are three-fold: 

\begin{itemize}
    \item Our main contribution is a pipeline for collecting ground truth data for a task of depth denoising. Specifically, we propose a method for temporal and spatial alignment of RBG-D sequences, removing the need in accurate checkerboard calibration and hardware clock synchronization. 
    \item We propose a data-efficient training method based on out-of-fold training paradigm. This approach enables us to train a convolutional LSTM using very limited amount of data.
\end{itemize}

\section{Related work}

Classic filtration-based approaches such as bilateral filtering~\cite{Tomasi1998bf} are powerful sensor-independent single-image approaches, yet are not applicable to high noise scenarios. Several modifications~\cite{Kopf07jbf,Zhang2014rgf,hui2014depth} use the corresponding color information to guide depth filtering process, but suffer from texture copying artifacts. Richardt~\etal~\cite{richardt2012coherent} perform depth upsampling for a depth stream using joint bilateral filtering for spatial filtering and optical flow for temporal filtering. Yuan \etal \cite{yuan2018temporal} stack 240 FPS color camera and 30 FPS depth camera and propose an algorithm to reconstruct intermediate depth maps and estimate scene flow. Similarly to us, they use a pair of RGB-D sensors to build an aligned dataset, yet their application is completely different. A number of works \cite{Han2013,Wu2014,OrEl2015,haefner2019tpami,haefner2018fight} use shape-from-shading techniques with shading extracted from an RGB image to optimize for a detailed depth map. However, these methods require precise alignment of depth and color modalities and rely on strong assumptions about object's materials and albedo. Several papers~\cite{son2016learning,guo2018tackling,Marco2017deepToF,Agresti2019MPIRemoval} tackle a challenging problem of denoising Time-of-Flight (ToF) data. Son~\etal~\cite{son2016learning} use high-precision structured
light sensor to capture groundtruth depth along with ToF camera measurements and learn a neural network to remove multipath distortions. In~\cite{shen2013layer,basso2018robust,herrera2012joint}, authors explicitly model the noise and explore parametric models for depth denoising and camera undistortion. In~\cite{Kaiser2017}, Kaiser~\etal fit primitives such as planes to depth data to perform denoising and hole filling.






Recently, \cite{hui2016depth,tang2019learning,li2019high,zuo2019residual} used deep convolutional neural networks to complete and upsample low-resolution depth maps using a corresponding high-resolution intensity image. Jeon~\etal~\cite{Jeon2018} use depth fusion~\cite{Dai2017bundlefusion} for scene reconstruction and synthesize groundtruth depth using the estimated surface and camera poses of each frame. They utilize a large dataset of indoor scenes~\cite{dai2017scannet} and propose a methodology for filtering wrong groundtruth depth patches. Having a dataset prepared, they learn a convolutional neural network for denoising. In~\cite{HyeokHyenKwon2015}, the authors use another reconstruction method~\cite{newcombe2011kinectfusion} to train their multi-scale dictionary of geometric primitives. Yan~\etal~\cite{Yan2018ddr} take a similar approach but focuses on human bodies. They use DoubleFusion~\cite{yu2018doublefusion} to fuse depth sequences of non-rigidly moving people. Although depth aggregation from multiple frames indeed leads to less noisy surfaces, fusion methods frequently fail to build consistent geometry and tracking errors lead to over-smoothing and loss of details. 

The method of Sterzentsenko~\etal~\cite{sterzentsenko2019self} is the most related to ours. Similarly to us, the authors use simultaneous recording with multiple well-synchronized Intel RealSense D415 sensors and train a deep network to perform denoising. In this paper we tackle a more complicated case, when the sensors can't be precisely synchronized, have different frame rate and intrinsic parameters. Moreover, rather than using a complex color re-projection loss as in~\cite{sterzentsenko2019self} we use a simple L1 loss and re-projected depth frame as ground truth. 

Also there is a proposed Noise2Noise framework~\cite{lehtinen2018noise2noise,Batson2019Noise2Self,Krull2019Noise2Void} reduces the need in clean groundtruth data yet relies on strong assumptions about the noise model such as spatial independence and zero mean value. Real-world sensors exhibit a complex noise model\cite{wasenmuller2016comparison} which limits the applicability of the mentioned approaches.     
\section{Overview}

Our self-supervised depth refinement approach starts with data collection. We create a dataset \( \{(S^{k}_{\Low}, S^{k}_{\High}){:}~k{\in}\intrange{1}{K}\} \) of $K$ paired sequences by simultaneously capturing streams $S^{k}_{\Low}$ and $S^{k}_{\High}$ from lower- and higher-quality sensors, respectively. In the following text, we omit the sequence index $k$ for notational clarity. Each sequence is defined by a set of measurements endowed with timestamps:
\begin{equation}\label{eqn:eqlabel}
\begin{alignedat}{3}
S_{\Low}  &= \{ \, (\Color^i_{\Low}, \, && \Depth^i_{\Low}, \, & & \Ts^i_{\Low}){:}~i\in\intrange{1}{N_\Low} \}\,,
\\
S_{\High} &= \{ \, (\Color^i_{\High}, \, && \Depth^i_{\High}, \, & & \Ts^i_{\High}){:}~i\in\intrange{1}{N_{\High}} \}\,.
\end{alignedat}
\end{equation}

Here, $\Color^i$, $\Depth^i$ stand for \(i\)'th color and depth images respectively, $\Ts^i$'s are corresponding timestamps, and $N$ is the sequence length. Subscripts $\Low$ and $\High$ stand for lower- and higher-quality sequences. Note, that timestamps $\Ts^i_{\High}$ and $\Ts^i_{\Low}$ are not directly comparable as the devices are not synchronized in time. 

\begin{figure}

\centering

\tikzstyle{decision} = [diamond, draw, fill=blue!20, 
    text width=4.5em, text badly centered, node distance=3cm, inner sep=0pt]
\tikzstyle{block} = [rectangle, draw, fill=blue!20, 
    text width=5em, text centered, rounded corners, minimum height=4em]
\tikzstyle{line} = [draw, -latex']
\tikzstyle{cloud} = [draw, ellipse,fill=red!20, node distance=3cm,
    minimum height=2em]
    
\begin{tikzpicture}[node distance = 3cm, auto]
    \node [block] (match) {Match frames (\ref{s:match})};
    \node [block, right of=match] (calibrate) {Find extrinsic (\ref{s:calibrate})};
    \node [block, right of=calibrate, node distance=3cm] (eval) {Evaluate loss (eq.~\ref{eq:spatial_loss})};
    \path [line] (match) -- (calibrate);
    \path [line] (calibrate) -- (eval);
    \path [line] (eval.north) |- ([yshift=0.4cm] eval.north) -- node  [above,midway] {For each possible time shift $\D$} ([yshift=0.4cm] match.north)  -| (match.north);
\end{tikzpicture}
\caption{Summary of the data processing pipeline. See~\cref{s:dataset} for details.}\label{fig:data_pipeline}
\end{figure}
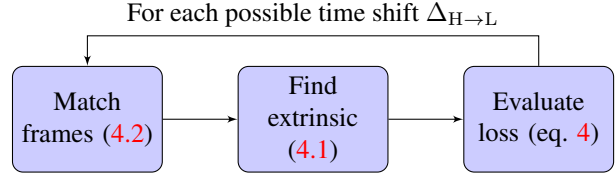

We then align the recorded sequences both in time and space. We assume that the sensors' clocks are out of sync and deviate from each other by an additive constant. For each sequence, we search for a shift $\D$, which would align the clocks of the two sensors~(\cref{s:match}). For each possible time shift $\D$, we extract indices of matching frames $\{(i, \Phi(i)): i \in [1\dots N_L]\}$ for $\Phi(i)$ defined as 
\begin{equation}
\Phi(i; \D) = \arg \min_j \| t^i_{\Low} - (t^j_{\High} + \D) \|\,.
\end{equation} 
We then search for an extrinsic matrix $\T$ to transform point clouds captured with the HQ sensor to the LQ camera coordinates by matching key points extracted from the RGB images $\{ \Color_\High^{\Phi(i)} \}$ and $\{ \Color_\Low^i \}$ (\cref{s:calibrate}). We project the transformed point clouds using the LQ camera projection matrix to get the corresponding ground truth for the LQ depth images. We choose the shift $\D$ such that it delivers the best spatial alignment in the extrinsic matching step. The pipeline is summarized in~\cref{fig:data_pipeline}. 

\tikzstyle{format} = [draw, thin, fill=blue!20]
\tikzstyle{medium} = [ellipse, draw, thin, fill=green!20, minimum height=2.5em]

Having a dataset of spatially and temporally aligned RGB-D frames, we interpret depth denoising as an image-to-image translation task and train a deep convolutional neural network to map RGB-D images coming from the LQ sensor to the reprojected HQ depth frames (\cref{s:training}).

We train a simple UNet-like neural network for single image denoising. To leverage the temporal information,  
we employ an out-of-fold prediction strategy~\cite{wolpert1992stacked} and train a ConvLSTM~\cite{xingjian2015convolutional} on top of the UNet's predictions. At test time, it takes a single forward pass of the UNet network to perform depth denoising. If sequential data is available, we use the second-level ConvLSTM to postprocess UNet predictions.

\section{Dataset}\label{s:dataset}

To create a dataset with aligned data for self-supervised learning, we fix two RGB-D sensors as rigidly as possible. In our setup (\cref{fig:teaser}~(a)) we used Kinect V2 (K2) and TrueDepth (TD) sensor embedded in iPhone X. We recorded $51$ simultaneous RGB-D sequences with different human actors, each about $30$ seconds long. We asked the actors first to make a turnaround in front of camera and then move freely for the remaining time. We used six different environments to maximize the diversity of the recorded data. The resolution of K2 and TD depth is \texttt{512x424} and \texttt{480x640} respectively.  

\subsection{Spatial alignment}\label{s:calibrate}

While in general it is possible to calibrate extrinsic parameters of the sensor pair using a checkerboard~\cite{zhang1999flexible}, we did not find such calibration robust enough for our scenario. In fact, while the checkerboard calibration worked well for the first sequence recorded after the calibration, the estimated extrinsic did not prove to be reliable for the following sequences because of physical interaction with iPhone, overheating, non-rigidity of K2 and our rig. Therefore, we perform in-the-wild extrinsic matrix calibration separately for each sequence pair using the available unsynchronized RGB-D data.

Having an estimate of the time shift $\D$ we extract the set of matched frames $\{(\Color^i_{\Low}, \Depth^i_{\Low}, \Color^{\Phi(i)}_{\High}, \Depth^{\Phi(i)}_{\High}):~i \}$. We then optimize for an extrinsic matrix $\T$ that would transform HQ sensor coordinate system into the LQ sensor system. For a point $p_{\High} = (x, y)$ on the image plane of HQ camera and the corresponding depth $\Depth^{\Phi(i)}_{\High}(p_{\High})$, the location  in 3D HQ camera space is defined as $P_{\High}(p_{\High}) = \pi^{-1}_{\High} ( p_{\High},\, \Depth^{\Phi(i)}_{\High}(p_{\High}))$ where $\pi^{-1}$ is an unprojection operator. The 2D position on the LQ frame is then determined as 
\begin{equation}\label{eq:reprojection}
  \hat p (p_{\High};\, \T) = \pi_{\Low} (\T P_{\High}(p_{\High}))\,,
\end{equation}

where $\pi_{\Low}$ is a projection function mapping 3D points in LQ camera space onto the camera plane with z-buffer. We compute RGB images $\hat \Color^{\Phi(i)}_{\High}$ by using~\ref{eq:reprojection} to reproject color frames $\Color^{\Phi(i)}_{\High}$ according to the current estimate of $\T$.

%

For each frame pair $(i, \Phi(i))$ we use the Superpoint detector~\cite{detone2018superpoint} to extract a set of 2D correspondences $ U_{i}$ from $\Color^i_{\Low}$ and $\hat \Color^{\Phi(i)}_{\High}$ and minimize the distance on the image plane between them with respect to the transform $\T$:
\begin{equation}\label{eq:spatial_loss}
  \mathcal{L}(\T) = \sum_{i=1}^{N_L} \sum_{ (p_{\Low}, p_{\High}) \in U_{i} } \| \hat p(p_{\High},\, \T) - p_{\Low}\|_2\,.
\end{equation}

We initialize the extrinsic matrix with identity transform $\T = I$ and parametrize the rotation component with quaternions. We use gradient descent for simplicity and flexibility. 

We found such iterative recomputation of the correspondences and gradient-based optimization to be robust to incorrect key points and occlusions.

 
 

\subsection{Temporal alignment}\label{s:match}

As it is impossible to enable hardware synchronization for the pair of the depth sensors we used, we implemented software timestamp synchronization and frame filtering as  postprocessing. Our alignment method builds a mapping $\Phi$ which maps an index $i$ of a LQ frame to an index $\Phi(i)$ of the corresponding HQ frame shot closely in time and tackles the following issues: 
\begin{itemize}
    \item Clocks of TD and K2 are not synchronized. That is, the two devices assign different timestamp to any point in time.
    \item The devices start recording at a different time.
    \item The devices have different frame rate. The frame rates are not constant. 
\end{itemize}

Importantly, we assume that the time coordinate system for the two sensors differs only by a constant shift $\D$ and the timestamps do not "drift" with time. That is, to transform an \textit{arbitrary} timestamp from HR sensor $\Ts_{\High}$ into LR sensor timestamp $\Ts_{\Low}$ we just need to add $\D$ to the former: 
$$
\Ts_{\Low} = \Ts_{\High} + \D\,.
$$

Note that we do not assume that the sensors have the same frame rate. The frame rate can vary over time for both sensors. We only make assumption about the relation of time coordinate systems.

For each pair of sequences we seek for a shift $\D$ that would align the timestamps of the two sensors so that a simple nearest neighbour search between the timestamps $\{ \Ts^i_{\Low}{:}~i\in\intrange{1}{N_{\Low}} \}$ and $\{ \Ts^j_{\High} + \D{:}~i\in\intrange{1}{N_{\High}} \}$ would give us the best mapping $\Phi$. We use binary search in the $[-60\text{ms}, 60\text{ms}]$ segment with $5\text{ms}$ step size to find the best $\D$ based on the correspondence loss defined in ~\cref{eq:spatial_loss}. 



\newlength{\step}
\setlength{\step}{0.8cm}

\newlength{\stepx}
\setlength{\stepx}{0.9cm}

\definecolor{mygray}{RGB}{200,200,200}
\tikzset{
    box1/.style={%
        rectangle,
        rounded rectangle,
        minimum height=0.7cm,
        minimum width=2cm
    },
}

\definecolor{mycolor}{RGB}{255,51,76}
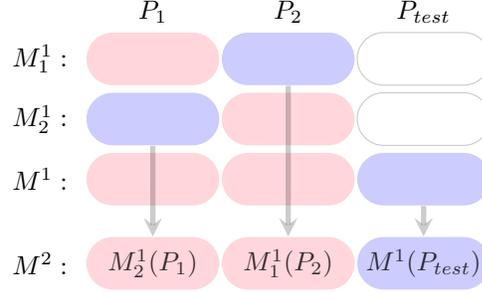
\begin{figure}
  \begin{tikzpicture}
    \node[box1, fill=mycolor, fill opacity=0.2, label={$P_1$}] (a11) at (0,0) {};
    \node[box1, fill=blue, fill opacity=0.2, label={$P_2$}] (a12) at (2\stepx,0.0) {};
    \node[box1, fill opacity=0.2, draw=mygray,  label={$P_{test}$}] (a13) at (4\stepx,0.0) {};
    
    \node[box1, fill=white, minimum width=0.2cm, fill opacity=1.0] (a13) at (-1.5,0.0) {$\U_1:$};
    
    \node[box1, fill=blue, fill opacity=0.2] (a21) at (0, -\step) {};
    \node[box1, fill=mycolor, fill opacity=0.2] (a22) at (2\stepx, -\step) {};
    \node[box1, fill opacity=0.2, draw=mygray] (a23) at (4\stepx, -\step) {};
    
    \node[box1, fill=white, minimum width=0.2cm, fill opacity=1.0] (a13) at (-1.5,-1\step) {$\U_2:$};
        
    \node[box1, fill=mycolor, fill opacity=0.2] (a31) at (0\stepx, -2\step) {};
    \node[box1, fill=mycolor, fill opacity=0.2] (a32) at (2\stepx, -2\step) {};
    \node[box1, fill=blue, fill opacity=0.2] (a33) at (4\stepx, -2\step) {};
    
    \node[box1, fill=white, minimum width=0.2cm, fill opacity=1.0] (a13) at (-1.5,-2\step) {$\U:$};
    
    \node[box1, fill=mycolor, text opacity=0.8, fill opacity=0.2] (a41) at (0\stepx, -3.4\step) {$\U_2(P_1)$};
    \node[box1, fill=mycolor, text opacity=0.8, fill opacity=0.2] (a42) at (2\stepx, -3.4\step) {$\U_1(P_2)$};
    \node[box1, fill=blue, text opacity=0.8, fill opacity=0.2] (a43) at (4\stepx, -3.4\step) {$\U(P_{test})$};
    
    \node[box1, fill=white, minimum width=0.2cm, fill opacity=1.0] (a13) at (-1.5,-3.4\step) {$\L:$};
    
    \draw [>=stealth,line width=2pt,opacity=0.2,->] (a12) edge (a42);
    \draw [>=stealth,line width=2pt,opacity=0.2,->] (a21) edge (a41);
    \draw [>=stealth,line width=2pt,opacity=0.2,->] (a33) edge (a43);
  \end{tikzpicture}
  \vspace*{2mm}\caption{Out-of-fold prediction scheme. We split all sequences into three groups $\{ P_1, P_2, P_{test} \} $. We then learn three models $\U_1$, $\U_2$, $\U$ on the group shown in red and get their predictions on the blue parts. For example, the first model is trained on the part $P_1$ and evaluated on the part $P_2$. We then use the predictions $\U_1(P_2)$ and $\U_2(P_1)$ to train a second-level model $\L$ while using $\U(P_{test})$ for its validation.}\label{fig:oof}
\end{figure}


With both extrinsic $\T$ and the frame mapping $\Phi$ found, for each index $i$ we compute depth and color images $\hat \Depth^{\Phi(i)}_{\High}, \hat \Color^{\Phi(i)}_{\High}$ by reprojecting the corresponding HQ depth and color images $\Depth^{\Phi(i)}_{\High}, \Color^{\Phi(i)}_{\High}$ to the LQ camera using~\cref{eq:reprojection}. Finally, we construct a sequence $\hat S_{\High} = \{\hat \Color^{\Phi(i)}_{\High}, \hat \Depth^{\Phi(i)}_{\High}, \Ts^i_{\Low} \}_{i=1}^{N_{\Low}}$ that corresponds to LQ sequence $S_{\Low}$.





\section{Training}\label{s:training}



To exploit temporal information available in the consecutive frames it is tempting to use a recurrent model. However, we did not succeed to directly train one due to limited amount of data. Instead, we utilize two-level training approach based on out-of-fold predictions approach~\cite{wolpert1992stacked} widely used for model ensembling. First, we train first-level UNet~\cite{ronnUnet}-like architecture to denoise depth on per-frame basis. As a second-level model we train a convolutional LSTM (ConvLSTM) to account for temporal correlations in the data.

To train the first-level model, we split all available sequences into three groups $P_1$, $P_2$, and $P_{\mathrm{test}}$. We use the sequences from the group $P_1$ to train a UNet $\U_1$ and the group $P_2$ to train a UNet $\U_2$ (\cref{fig:oof}). We use those models to get the predictions for an \textit{unseen} group of the sequences (e.g. a model trained on the part $P_1$  is used to predict depth for the part $P_2$ and vice versa). Additionally, we train a UNet $\U$ on both parts $P_1$ and $P_2$ and use it to get predictions for the testing sequences $P_{\mathrm{test}}$.

To this end, we constructed a training set $\{ \U_1(P_2), \U_2(P_1) \}$ and a testing set $\U(P_{\mathrm{test}})$ for a second-level model that consist of first-level models' predictions. Note, that by employing out-of-fold strategy we create an \textit{unbiased} dataset of the same size as the original dataset, that we now can use to train a second-level model.

We use a ConvLSTM $\L$ as a second-level predictor. We found it efficient at capturing temporal correlations in data and refining single-image depth denoising result of first-level's UNet $\U$. 

To account for synchronization errors, we first use frames aligned within $15$ms to train both first- and second-level models and then finetune ConvLSTM using only $5$ms distant frames.  

\paragraph{Losses.}

While we could employ photometric reprojection losses as in~\cite{sterzentsenko2019self}, the availability of the groundtruth depth images allows us to use a simple per-pixel loss between the reprojected K2 depth frames $\hat \Depth_\High^{\Phi(i)}$ and model's output $\Depth_{\mathrm{pred}}^i$:

\begin{equation}
\mathcal{L}(\Depth_{\mathrm{pred}}^{i}, \hat \Depth_\High^{\Phi(i)}) = \| (\Depth_{\mathrm{pred}}^{i} - \hat \Depth_\High^{\Phi(i)}) \cdot m^{\Phi(i)} \cdot m_{\mathrm{seg}}^i \|_1\,.
\end{equation}

Here, $m$ is a binary mask that indicates the presence of the depth information for each pixel of $\hat \Depth_\High^{\Phi(i)}$ and $m_{\mathrm{seg}}^i$ is person segmentation mask precomputed based on the color image $\Color_{\Low}^i$ using~\cite{gong2019graphonomy}. The model predicts a dense depth image while the reprojected HQ depth frames $\hat \Depth_\High^{\Phi(i)}$ are sparse as the resolution of the K2 sensor is lower that the resolution of TD camera.
 

\section{Testing}

At the testing time, given an input RGB-D frame coming from the LQ sensor, we feed it to the first-level UNet $\U$ to estimate a denoised depth image. Being trained on humans, the network's predictions are only reliable at the pixels corresponding to human bodies. Therefore, we multiply the predicted depth image pixel-wise by person segmentation mask. While the basic model $\U$ can be used both in single-image and multi-image scenarios, when sequential data is available, we can use the full model with the second-level ConvLSTM $\L$. We aggregate the predictions of the first-level model $\U$ and feed them one by one to $\L$ to produce a final denoised depth map for each frame of the LQ sequence. 

We note that even though our training pipeline is not end-to-end, the testing pipeline can be implemented as such by merging first- and second-level models in a single predictor. Yet, we found it to be technically easier to apply models $\U$ and $\L$ one after another saving intermediate predictions to a hard drive.




\section{Experiments}

\begin{table*}
\centering
%
%


\caption{Quantitative comparison of depth denoising approaches. Our method reaches the lowest average MSE error between predicted depth frames and groundtruth depth frames. }\label{tab:comparison}
\vspace*{2mm}

\setlength{\tabcolsep}{5pt}
\begin{tabular}{@{}lccccccccc@{}}\toprule
     & Raw 
     & {\mutlirow{\small JBF}{\small \cite{Kopf07jbf}}}
     & \small{\mutlirow{RGF}{\cite{Zhang2014rgf}}}
     & \small{\mutlirow{BF}{\cite{Tomasi1998bf}}}
     & \small{\mutlirow{DDD}{\cite{sterzentsenko2019self}}}
     & \small{\mutlirow{DDRNet}{\cite{Yan2018ddr}}}
     & \small{\mutlirow{DRR}{\cite{Jeon2018}}}
     & \small{\mutlirow{Ours}{(basic)}}
     & \small{\mutlirow{Ours}{(LSTM)}}
     \\ \midrule
    {MSE }{(mm)} 
    & $57.22$  
    & $49.69$ 
    & $49.49$  
    & $56.39$ 
    & $84.07$  
    & $57.08$
    & $53.32$
    & $31.61$
    & $\bf{21.02}$
\end{tabular}

\end{table*}

\begin{figure*}
    \centering
    \newlength{\mdd}
    \setlength{\mdd}{1.4cm}
    \newlength{\mddb}
    \setlength{\mddb}{1.1cm}
    \newlength{\mddc}
    \setlength{\mddc}{1.3cm}
    \newlength{\mddd}
    \setlength{\mddd}{1.5cm}
    \centering\resizebox{0.9\linewidth}{!}{
    \begin{tabular}{ccccc}
     \includegraphics[width=\mdd]{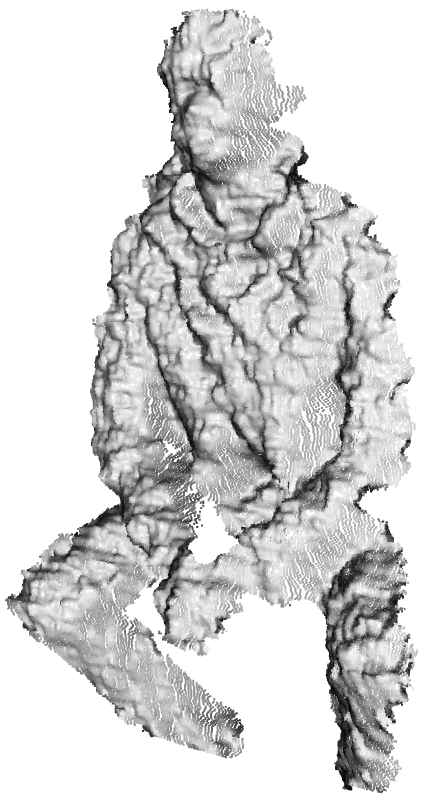}&
    \includegraphics[width=\mdd]{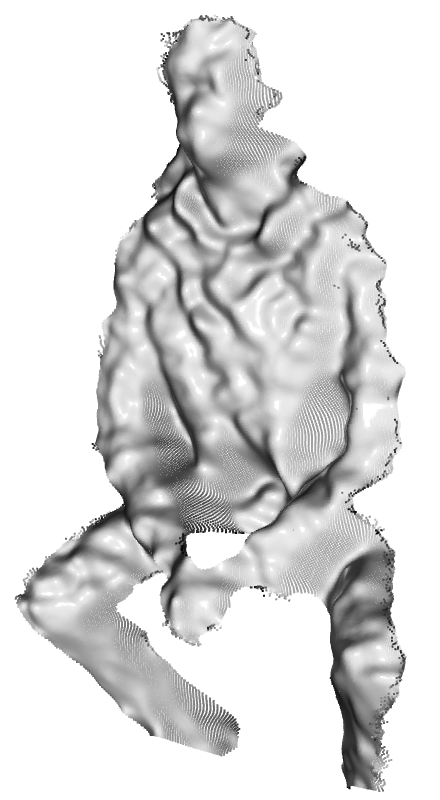} & 
    \includegraphics[width=\mdd]{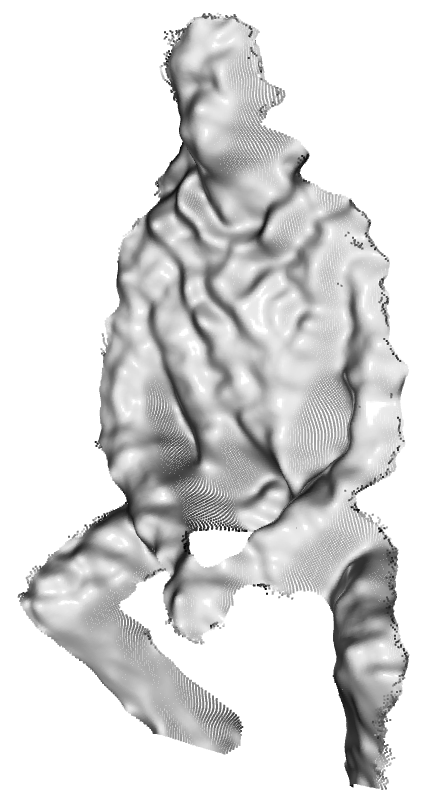} &
    \includegraphics[width=\mdd]{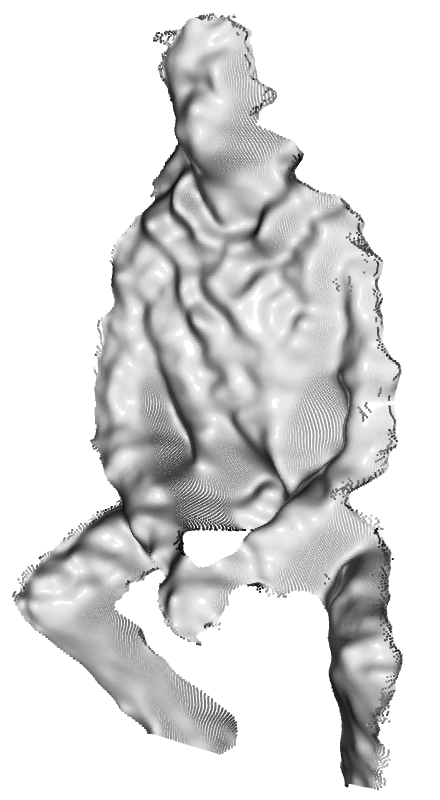}&
    \includegraphics[width=\mdd]{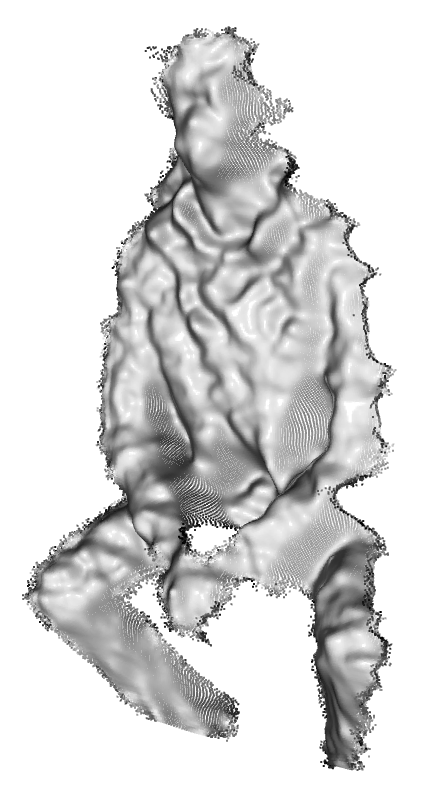} 
    \vspace*{-1mm}
    \\ 
    \scriptsize{(a) Input (TD)} & \scriptsize{(b) RGF} & \scriptsize{(c) JBF} & \scriptsize{(d) BF} & \scriptsize{(e) DDD} \\ 
    \includegraphics[width=\mdd]{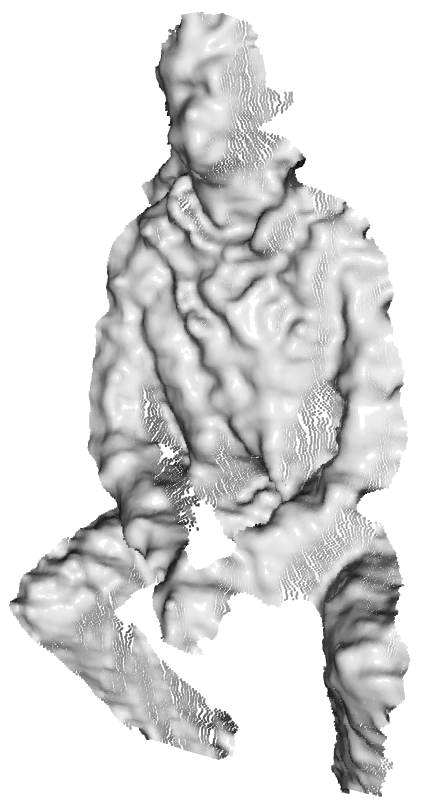} &
    \includegraphics[width=\mdd]{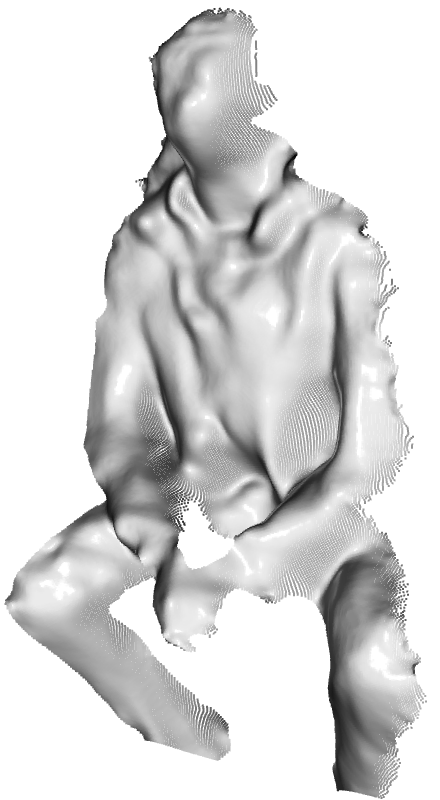}  &
    \includegraphics[width=\mdd]{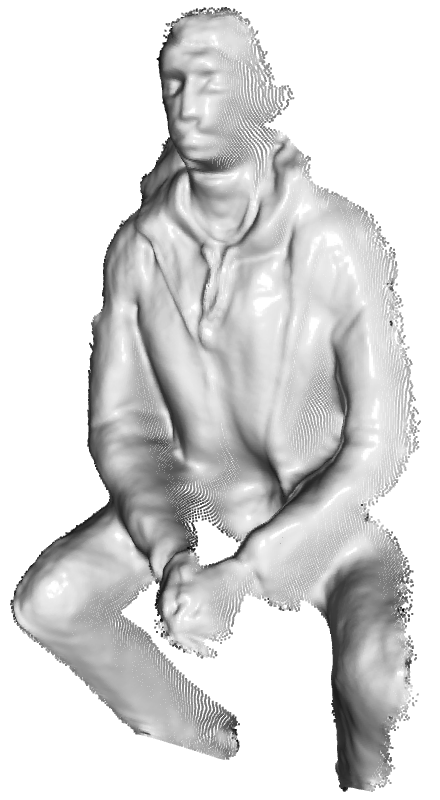}&
    \includegraphics[width=\mdd]{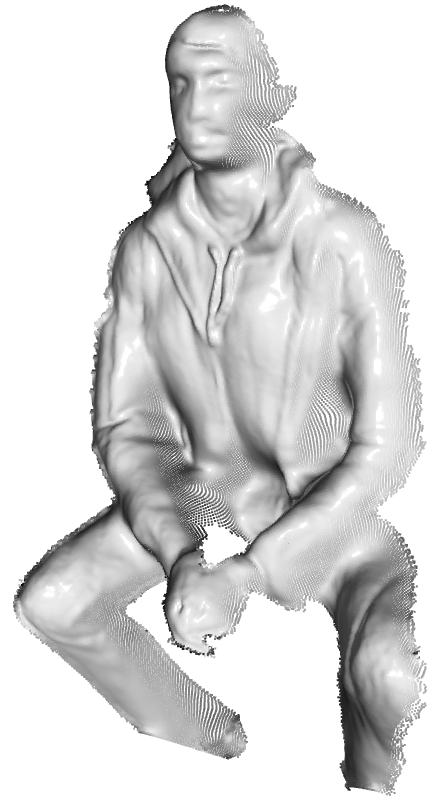}& 
    \includegraphics[width=\mdd]{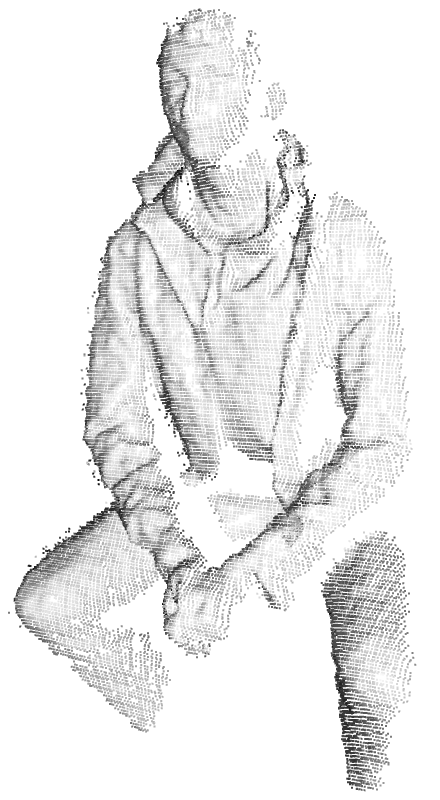} 
    \vspace*{-1mm}
     \\
    \scriptsize{(f) DDRNet}  & \scriptsize{(g) DRR} & \scriptsize{(h) Ours (basic)}  & \scriptsize{(j) Ours (full)} & \scriptsize{(k) K2}
    \\
    
    \end{tabular} }
    \caption{Qualitative comparison of depth denoising methods. Please see a video in supplementary materials.}\label{fig:comparison_lq}

\vspace{3mm} 
\centering
\centering\resizebox{0.93\linewidth}{!}{
\begin{tabular}{cccccc}
    \centering

    \includegraphics[width=1.4cm]{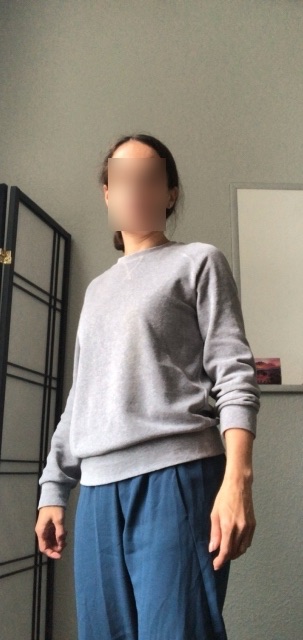} &
    
    \includegraphics[width=\mddc]{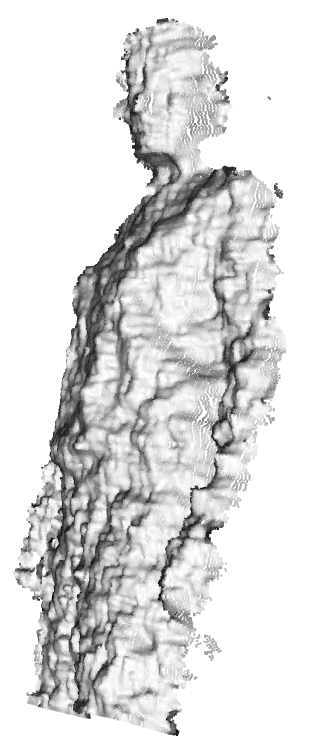} &
    
    \includegraphics[width=\mddc]{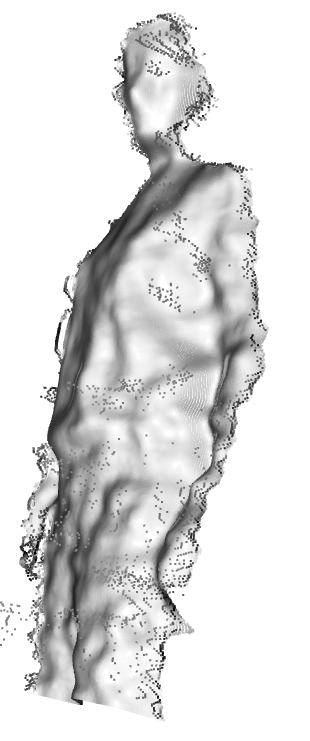} &
    
    \includegraphics[width=\mddc]{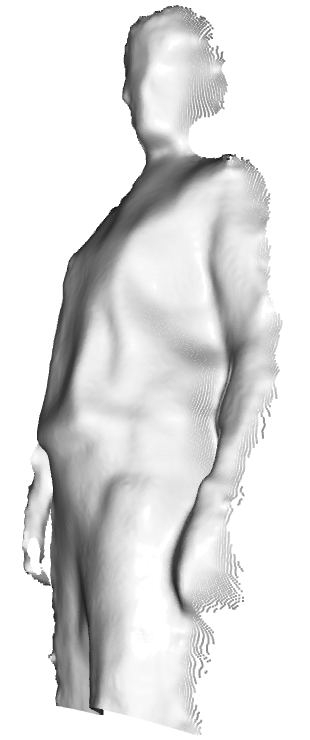} &
    
    \includegraphics[width=\mddc]{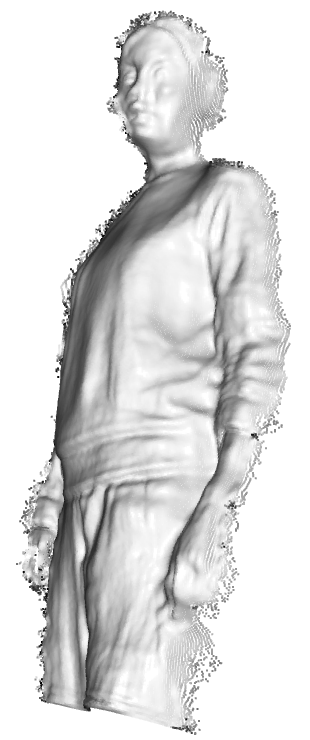}
    
    \vspace*{-1mm}
    \\
    \includegraphics[width=\mddd]{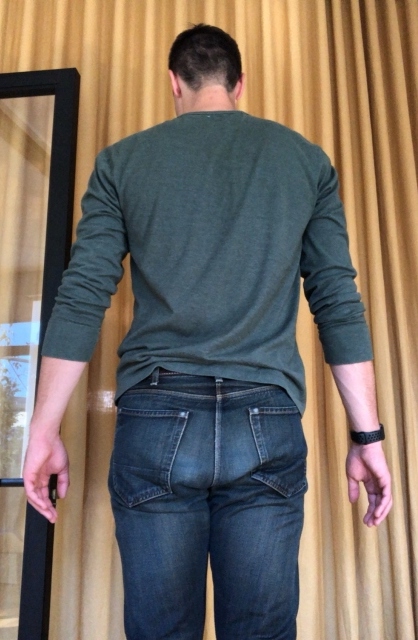}&
    
    \includegraphics[width=\mddd]{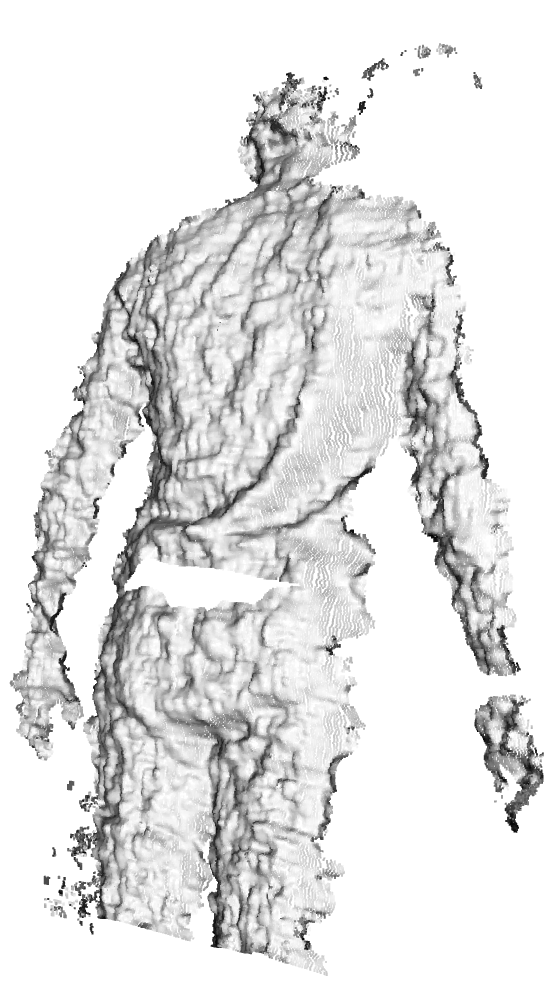} &
    
    \includegraphics[width=\mddd]{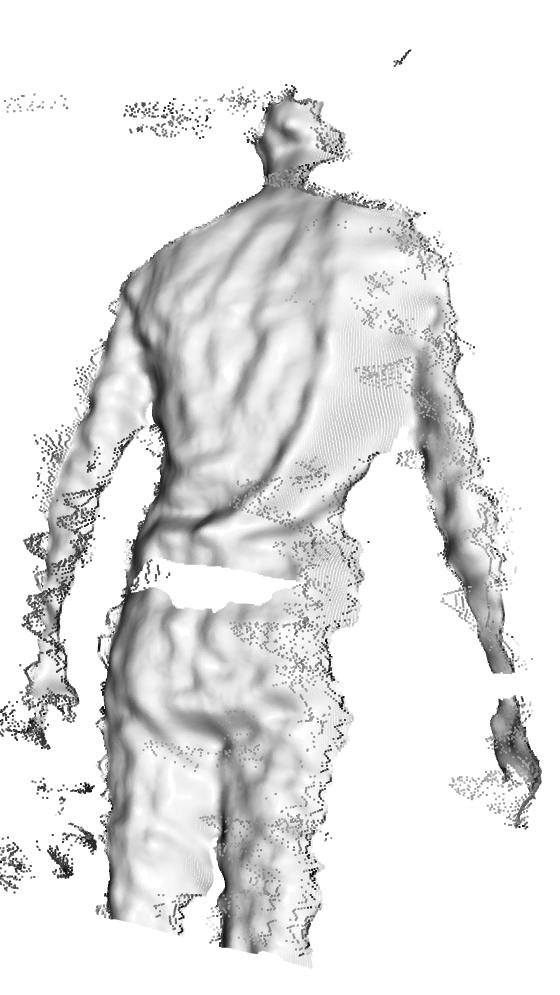} &
    
    \includegraphics[width=\mddd]{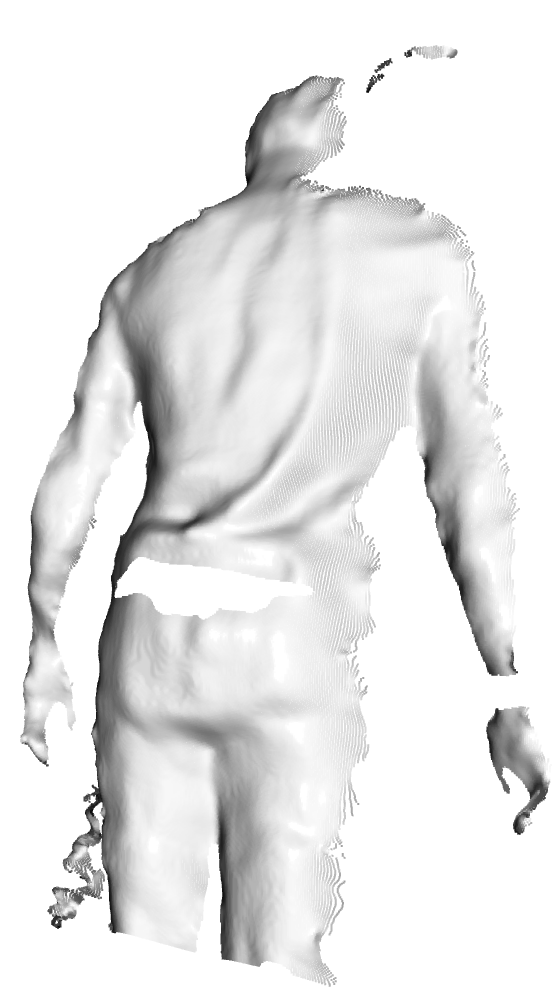} &
    
    \includegraphics[width=\mddd]{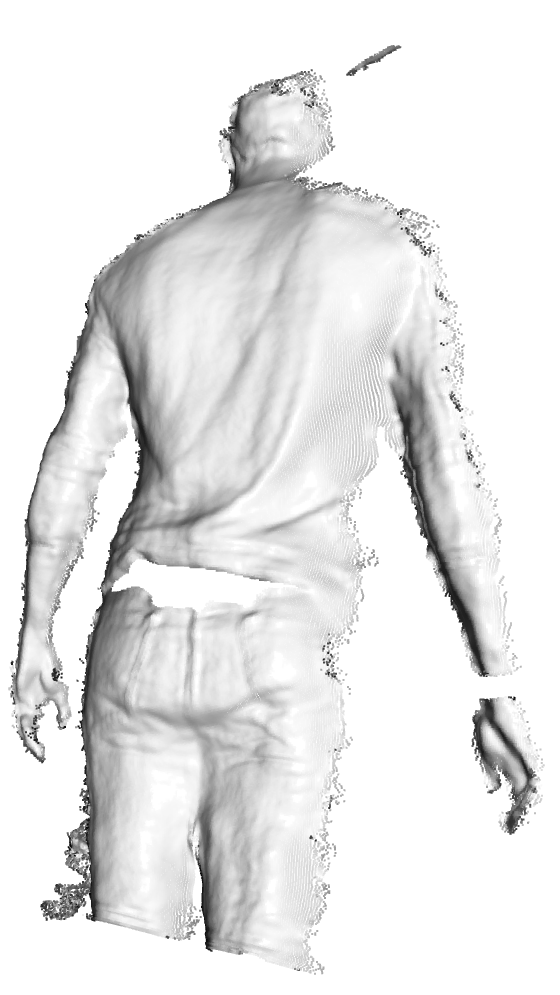} &
    \vspace*{-1mm}
    \\
    \scriptsize{(a) Color} & \scriptsize{(b) TrueDepth} &  \scriptsize{(c) RGF}  & \scriptsize{(d) DRR} & \scriptsize{ (e) Ours(basic)}
    \vspace*{-1mm}
    
\end{tabular}}
    \caption{Our model learned in a self-supervised way takes color and depth data coming from the sensor of an iPhone X as input (a, b) and produces a denoised and refined depth (e). Our method works significantly better than filtering-based (c) and other data-driven approaches (d). We intentionally visualize depth slightly rotated in 3D to highlight the differences between the methods. For more examples please refer to supplementary materials.}\label{fig:small_compar}
\end{figure*}

In this section, we experimentally validate the proposed method. We use four 30 second test sequences with actors and environments never seen during training. For filtering-based methods Bilateral Filter (BF~\cite{Tomasi1998bf}), Rolling Guidance Filter (RGF~\cite{Zhang2014rgf}),  Joint Bilateral Filter (JBF~\cite{Kopf07jbf}) we use their OpenCV~\cite{bradski2008learning} implementation and pick the parameters, that minimize mean squared error with respect to ground truth K2 depth images. We used the code provided by authors to evaluate DDRNet~\cite{Yan2018ddr}, DRR~\cite{Jeon2018}, DDD~\cite{sterzentsenko2019self}. Qualitative comparison is provided in~\cref{fig:comparison_lq} and quantitative results are summarized in~\cref{tab:comparison}.
   
Please note that DDD, DDRNet, DRR are trained on different datasets and it is not possible to retrain them using our data for various reasons discussed below.

\textbf{DDD.} Sterzentsenko~\etal explicitly claim DDD is sensor-agnostic: “[DDD] maintains its performance when denoising depth maps captured from other sensors.” We cannot train DDD on our data as DDD requires the sensors placed at a sufficient distance (DDD uses reprojection loss). 

\begin{figure*}
    \centering
    \centering\resizebox{1\linewidth}{!}{
    \begin{tabular}{@{}cccccccccc@{}}
    \includegraphics[width=1.5cm]{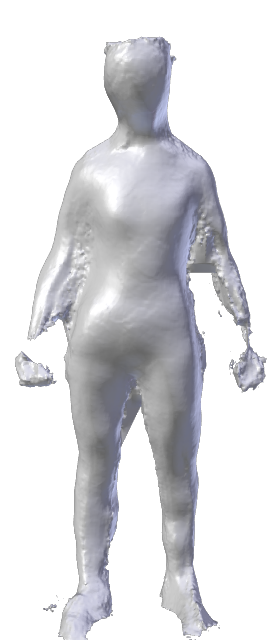} &
    \hspace{-3mm}\includegraphics[width=1.2cm]{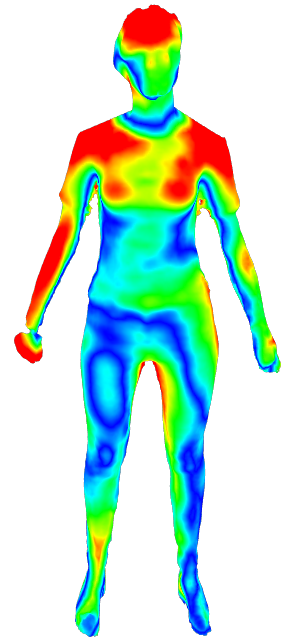} &
    \includegraphics[width=1.5cm]{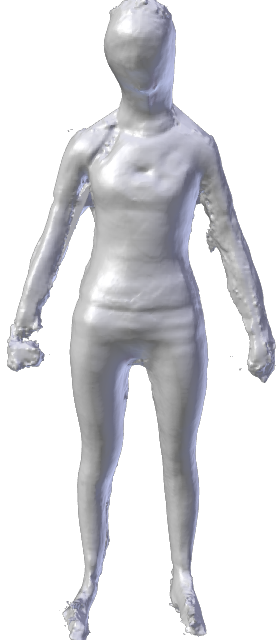}& \hspace{-5mm}\includegraphics[width=1.2cm]{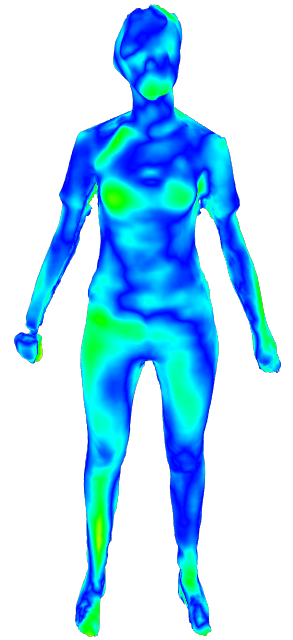} &
    \includegraphics[width=1.5cm]{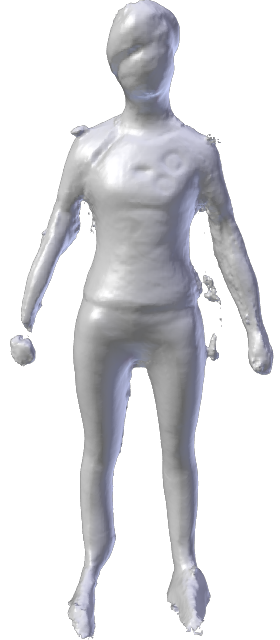} & \hspace{-5mm}\includegraphics[width=1.2cm]{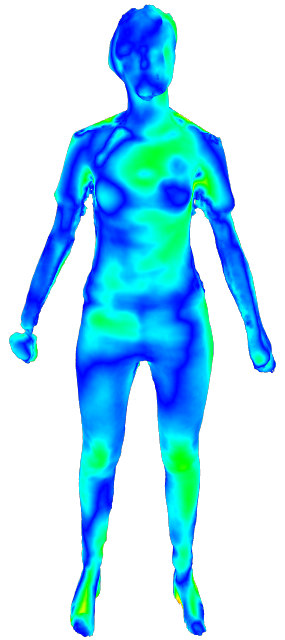} &
    \hspace{3mm}\includegraphics[width=1.9cm]{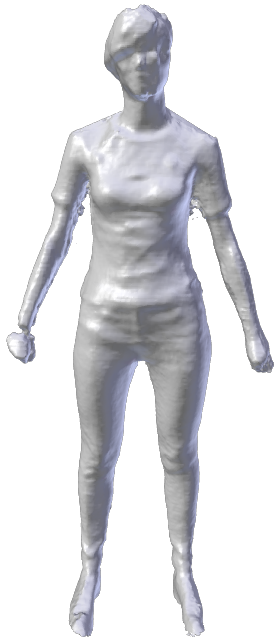} \\
    \scriptsize{(a) TD} & & 
    \scriptsize{(b) Ours (basic)} & & 
    \scriptsize{(c) Ours (full)} & &
    \scriptsize{(e) K2} 
    
    \end{tabular}}
    \caption{Canonical models built by DoubleFusion for different input data. DoubleFusion is designed to work with K2 input (e) and cannot handle noisy TD depth (a). Our method greatly improves the result of DoubleFusion (b, c) enabling its usage in mobile applications. We compare K2 mesh (e) with meshes (a, b c) by computing the distance between the surfaces and visualize the result as heatmaps. Blue color corresponds to zero error while red corresponds to 44mm error. }\label{fig:double_fusion}
\end{figure*}

\textbf{DDRNet.} DDRNet is traind on the data gathered with K2 sensor. Note, that it is very challenging to retrain DDRNet on TD data as the TD depth is usually too noisy to perform 3D reconstruction reliably. DDRNet uses Double fusion to create ground truth data. An example of DoubleFusion run on TD data is presented at~\cref{fig:double_fusion}~(a). It is obvious that the quality of the reconstructed mesh is insufficient to be used as ground truth.

\textbf{DRR.} DRR is trained on K2 data of static scenes. For the experiments, we used a model provided by authors. Our dataset consists of moving people and Kinect Fusion cannot be used to reconstruct the scene. Even for static scenes, it is still very challenging to run Kinect Fusion on TD data as TD data is too noisy.

Out of all filtering-based methods we find Rolling Guidance Filter to work best. Unfortunately, RGF is only efficient at removing high-frequency noise but not able to recover missing details. DDRNet, being trained on Kinect V2 data, also fails to adapt to the high noise magnitude. DRR network of Jeon~\etal is mostly trained on planar surfaces and thus over-smooths the input depth. While Sterzentsenko~\etal claim that DDD is sensor-agnostic, we did not manage to get good results with this method. 

The proposed method achieves $21$mm mean squared error which is lower than for the baselines. 

Finally, we compare our basic single-image denoiser $\U$ and the full model $\L$ with ConvLSTM on top of the predictions of basic model. We found that the full model benefits from aggregating depth measurements from subsequent frames and outperforms the single-image $\U$ by a huge margin.

\paragraph{Dynamic body reconstruction.}
Next, we use the proposed models for human body reconstruction and tracking application. We run DoubleFusion~\cite{yu2018doublefusion} on sequences from the validation set. We apply it to raw TD data, raw K2 data, and to the data obtained by applying the models $\U$ and $\L$ to the TD data. The qualitative comparison of the reconstructed canonical models for one of the sequences is given in~\cref{fig:double_fusion}. We compute average distance between the surfaces created using TD depth and K2 depth and get $13$mm for raw TD depth, $5.9$mm for the depth denoised with $\U$ and $5.7$mm for the depth denoised with $\L$. Clearly, denoising with our method significantly improves DoubleFusion reconstruction if compared to the result based on the raw TD data and significantly reduces the gap between TD and K2. 

\paragraph{Static body reconstruction.}
We use InfiniTAM~\cite{prisacariu2017infinitam} framework to assess an effect of our denoising procedure on the quality of the resultant mesh in a KinectFusion reconstruction pipeline. In this experiment, we first record a frame sequence, denoise the depth and then run the depth fusion algorithm with the denoised depth stream as input. We observe that in the cases of slowly-moving camera with high frame rate when a lot of frames can be accumulated, an accurate mesh can be produced even with the low-quality TD stream with objects at relatively large distances. In reality, the number of frames may be limited both by the recording conditions and by the available processing time. Thus, we focus on the setting when only a small number of frames available for reconstruction.

\begin{figure*}
    \centering\resizebox{.85\linewidth}{!}{
    \begin{subfigure}[b]{0.05\linewidth}
        \centering
        \includegraphics[width=\linewidth]{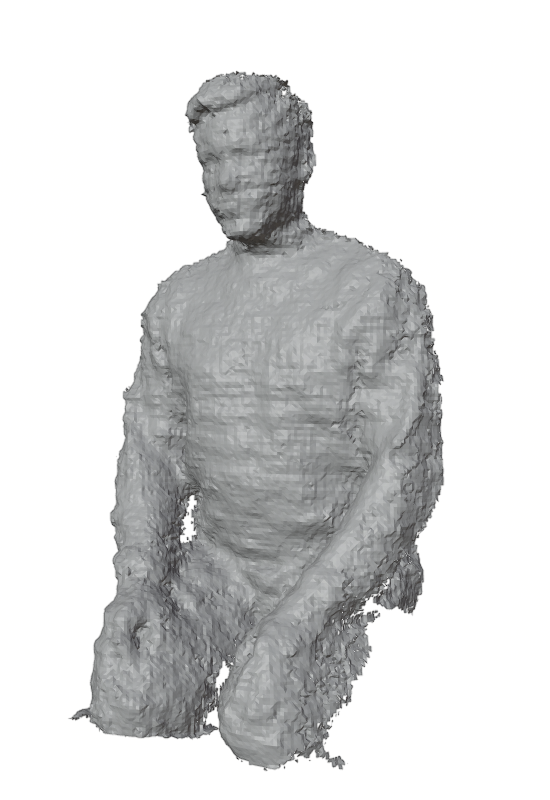}
    \end{subfigure}
    \begin{subfigure}[b]{0.05\linewidth}
        \centering
        \includegraphics[width=\linewidth]{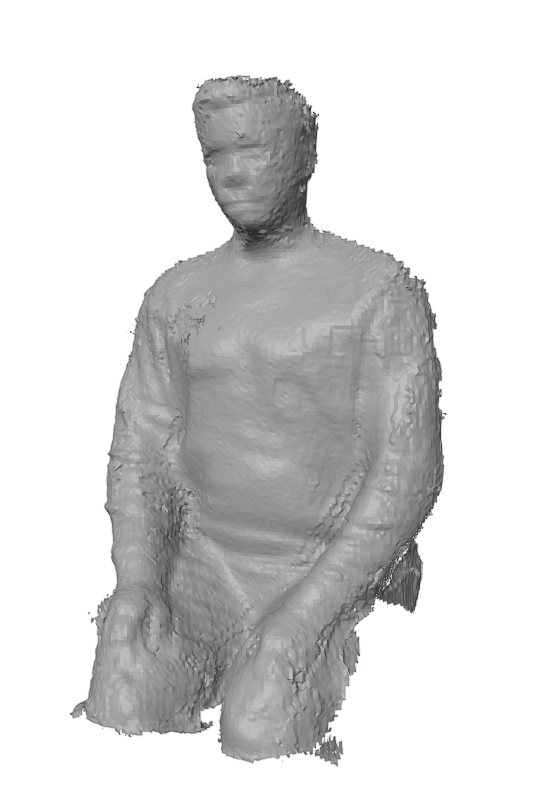}
    \end{subfigure}
    \begin{subfigure}[b]{0.05\linewidth}
        \centering
        \includegraphics[width=\linewidth]{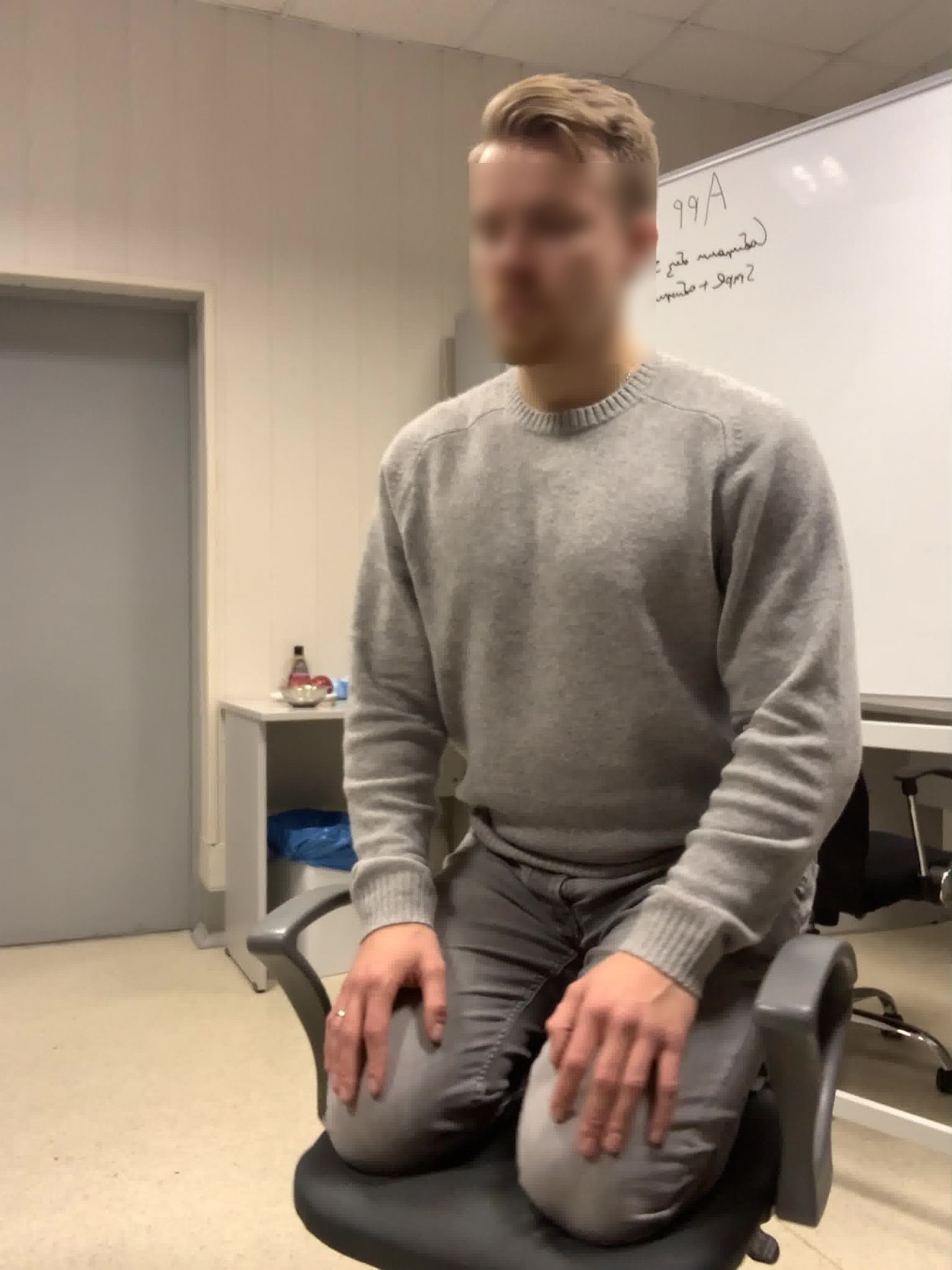}
    \end{subfigure}
    }%
    \caption{3D reconstruction using KinectFusion framework based on the raw depth stream from TD sensor (left) and based on the output of our denoising algorithm (center). The visual outlook of the scene is shown on the right. With denoised input, the quality of the produced model is higher and the level of noise is greatly reduced.}
    \label{fig:infinitam}
\end{figure*}

We fixed an LQ sensor statically and recorded an RGB-D stream of person rotating in the office chair.\footnote{This process is equivalent to moving camera around the person but does not require disassembling of our installation.} We removed the background using semantic segmentation mask and depth thresholding. During the recording of the video our test subject was rotated in the 90 degree arc, centered in the front-faced position. We picked seven evenly-spaced frames and run KinectFusion implemented via InfiniTAM. The resultant meshes are shown in~\cref{fig:infinitam}. Our method provides more details in each frame resulting in a higher detailed reconstructed geometry.

\subsection{Technical details}\label{s:technical_details}

In this section we provide technical details necessary to reproduce the results of the proposed method. 

\textbf{Network architecture.} We use a UNet with four downsampling operations. In addition to standard UNet architecture, we used skip connections between the input layer and each layer of the encoder.  

We used a ConvLSTM model similar to \cite{lstm1,lstm2}. We base it on UNet's architecture and add ConvLSTM layer after  every convolutional layer in the encoder and decoder. We train LSTM on the sequences of size four.


\textbf{Training and testing.} Both first-level and second-level models take color and depth images $(\Color^i_\Low, \Depth^i_\Low)$ along with person segmentation mask precomputed using~\cite{gong2019graphonomy} and XY ``meshgrid"~\cite{liu2018intriguing,ulyanov2018deep} as input. We concatenate the mentioned maps along the feature dimension.

We found that the pixel location information coming with "meshgrid" channels helps the network to better fit the data. Yet, we realize that this fact can also indicate about an error in the extrinsic/intrinsic calibration. In fact, while we use color information to find the transformation that would align sensor's coordinate systems, we heavily rely on the intrinsic and color-to-depth extrinsic calibration of \textit{each} device which we precompute once. With insufficiently accurate intra-sensor calibration, the photometric consensus between LQ color image $C^i_{\Low}$ and the reprojected HQ color image $\hat C^j_{\High}$ does not lead to the alignment between the pointclouds corresponding to the depth images $D^j_{\Low}$ and $\hat D^i_{\High}$  (e.g. LQ point cloud is always slightly rotated with respect to the HQ point cloud).   


We train UNet for 5 hours with batch size of 48 using four Titan V100 GPU's. The training of the second-level ConvLSTM model takes 2 days using the same hardware. We use Adam optimizer~\cite{adam} with a scheduler that lowers the learning rate when the loss stop to decrease. We use PyTorch~\cite{pytorch} in all our experiments.

\section{Conclusion and discussion}

In this paper we proposed a pipeline for self-supervised depth denoising and refinement. We overcome the lack of groundtruth data for data-driven depth denoising of low-quality depth sensors by employing a higher quality sensor and collecting simultaneous recordings. We automatically extract frames captured close in time and do not require the sensor pair to be precalibrated. We formulate the task of depth denoising as image-to-image translation and use a simple L2 loss between the depth images. While our method is easy to implement, it has a number of limitations. Our method is limited by the applicability of the depth sensors, e.g. Kinect V2 can't be used in direct sunlight setting. We also sometimes observe color ”leakage” (\cref{fig:comparison_lq} (j)) and imperfect temporal consistency (cf. video in supplementary material). Finally, being trained for a particular sensor types our models cannot be applied to the depth images captured with a another sensors.  

\clearpage
\section{Supplementary material}
\begin{figure*}
    \centering
    \newlength{\mdda}
    \setlength{\mdda}{1.cm}
    \newlength{\mddab}
    \setlength{\mddab}{0.8cm}
    \newlength{\mddac}
    \setlength{\mddac}{1.0cm}
    \newlength{\mddad}
    \setlength{\mddad}{1.2cm}
    \centering\resizebox{0.9\linewidth}{!}{
    \begin{tabular}{ccccc}
     \includegraphics[width=\mdda]{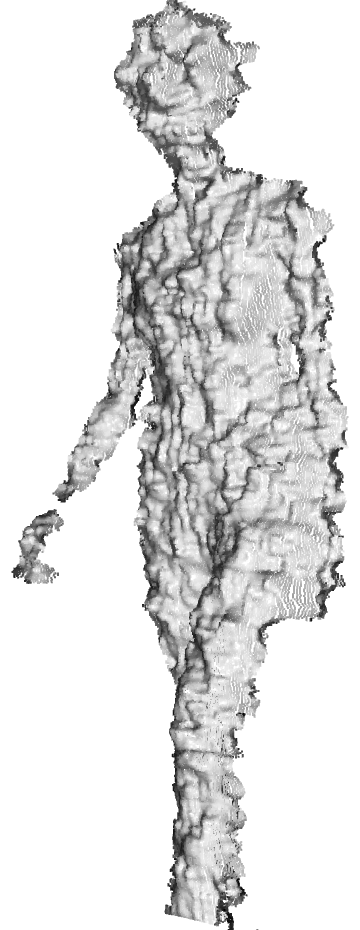}&
    \includegraphics[width=\mdda]{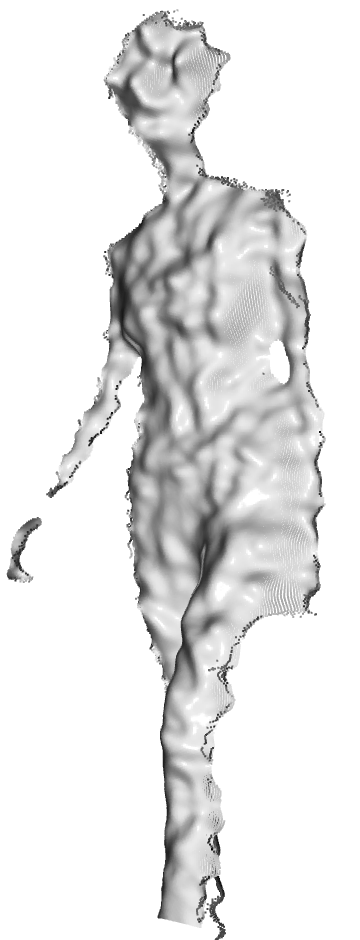} & 
    \includegraphics[width=\mdda]{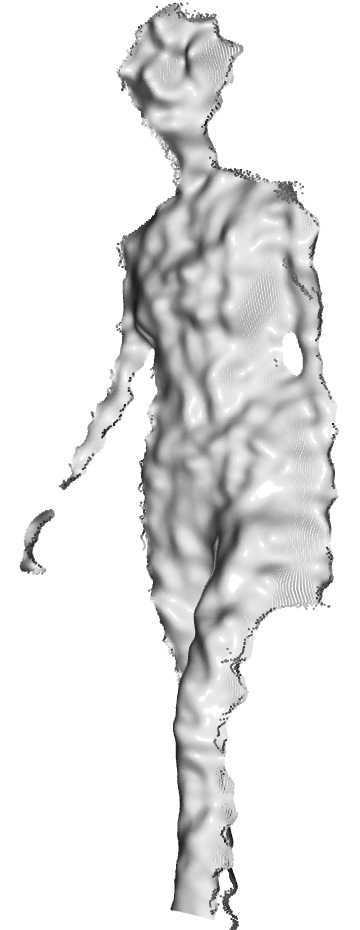} &
    \includegraphics[width=\mdda]{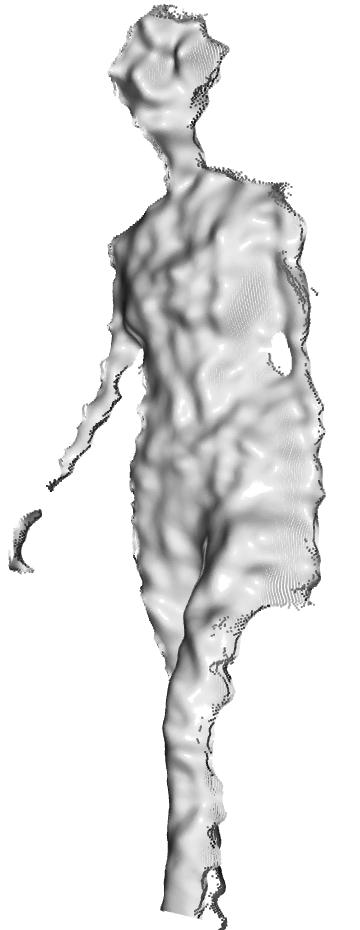}&
    \includegraphics[width=\mdda]{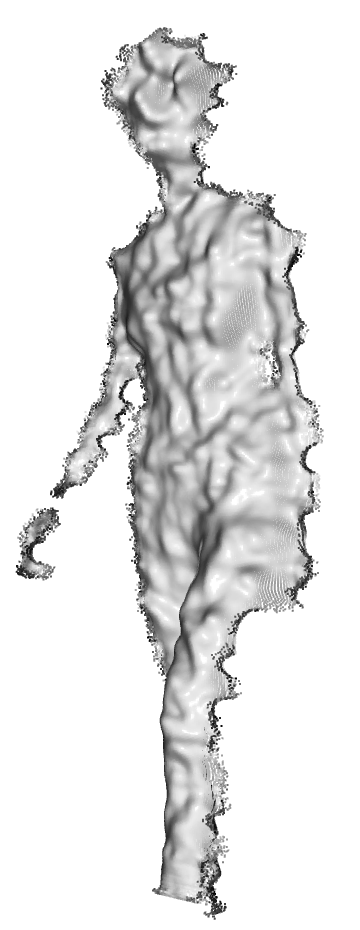} 
    \vspace*{-1mm}
    \\ 
    \scriptsize{(a) Input (TD)} & \scriptsize{(b) RGF} & \scriptsize{(c) JBF} & \scriptsize{(d) BF} & \scriptsize{(e) DDD} \\ 
    \includegraphics[width=\mdda]{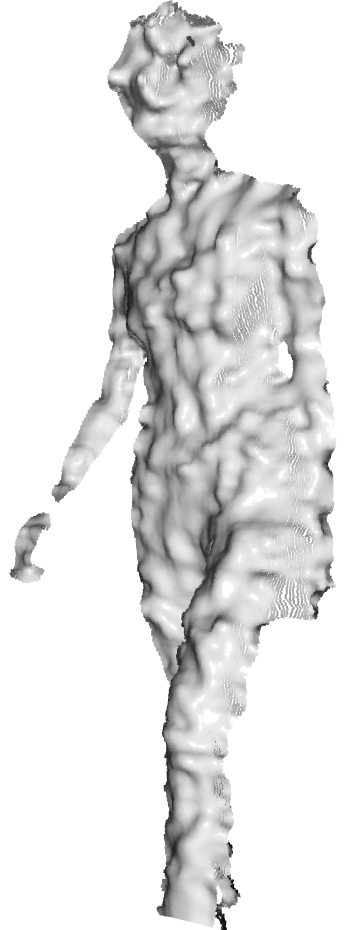} &
    \includegraphics[width=\mdda]{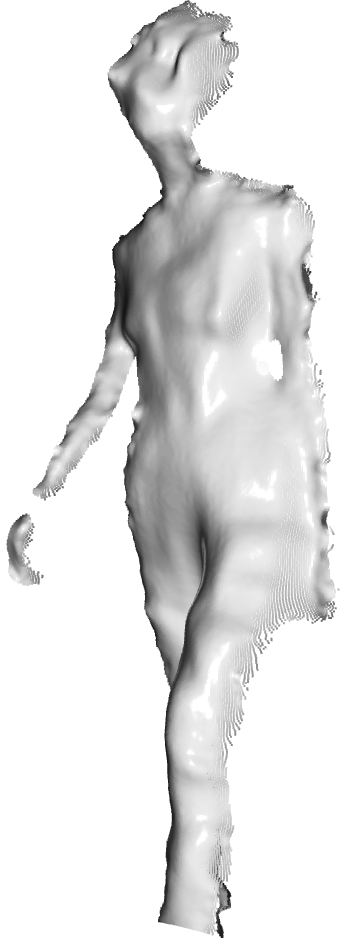}  &
    \includegraphics[width=\mdda]{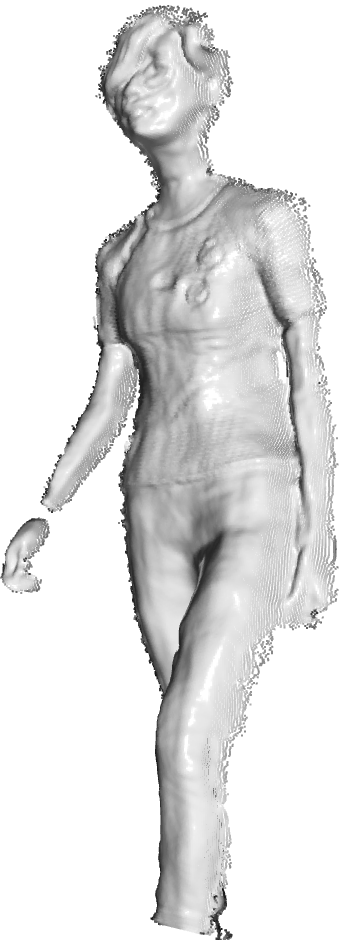}&
    \includegraphics[width=\mdda]{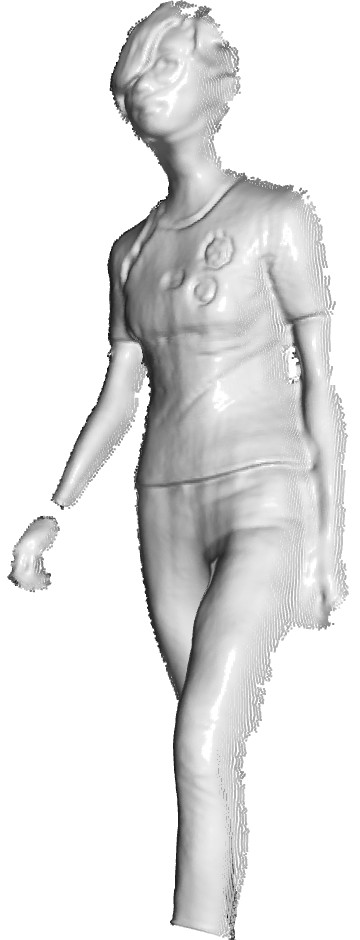}& 
    \includegraphics[width=\mdda]{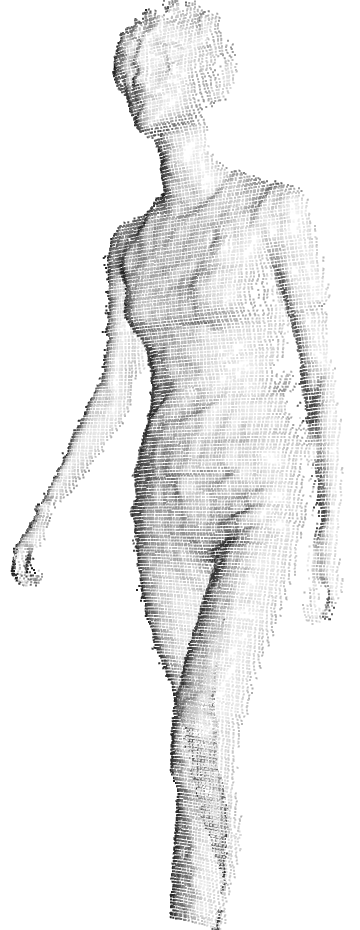} 
    \vspace*{-1mm}
     \\
    \scriptsize{(f) DDRNet}  & \scriptsize{(g) DRR} & \scriptsize{(h) Ours (basic)}  & \scriptsize{(j) Ours (full)} & \scriptsize{(k) K2}
    \\
    
    \end{tabular} }
    \caption{Qualitative comparison of depth denoising methods. Please see videos in addition to this PDF file in supplementary materials.}\label{fig:supmat_big_compar}

    \centering
    \newlength{\mddaa}
    \setlength{\mddaa}{1.1cm}
    \centering\resizebox{0.95\linewidth}{!}{
    \begin{tabular}{@{}ccccccc@{}}
    
    \includegraphics[width=1.4cm]{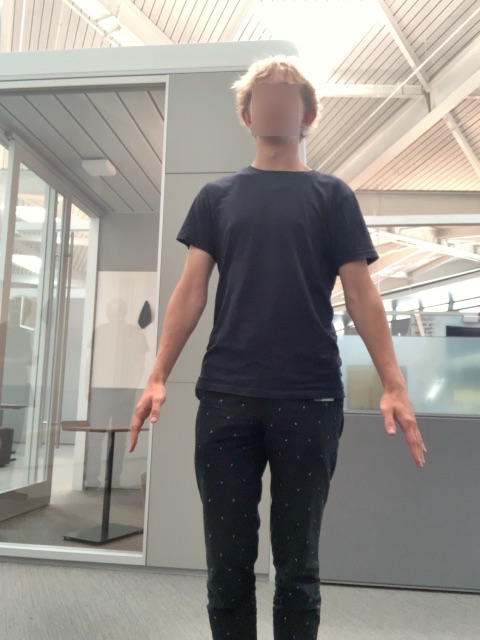} &
    \includegraphics[width=\mddaa]{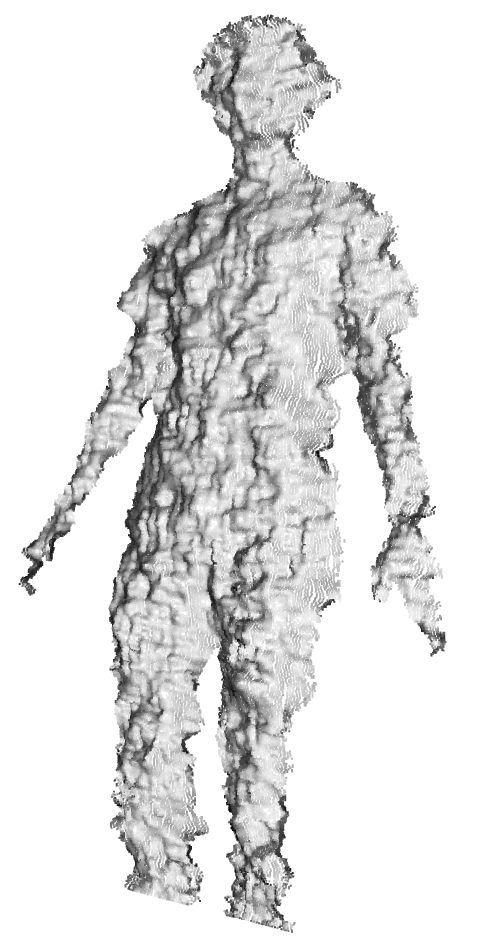} & 
    \includegraphics[width=\mddaa]{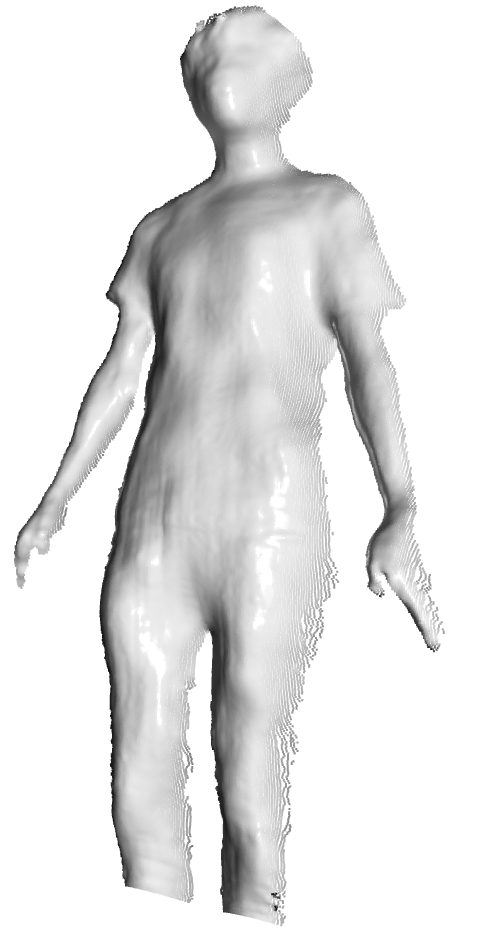} &
    
    \includegraphics[width=1.4cm]{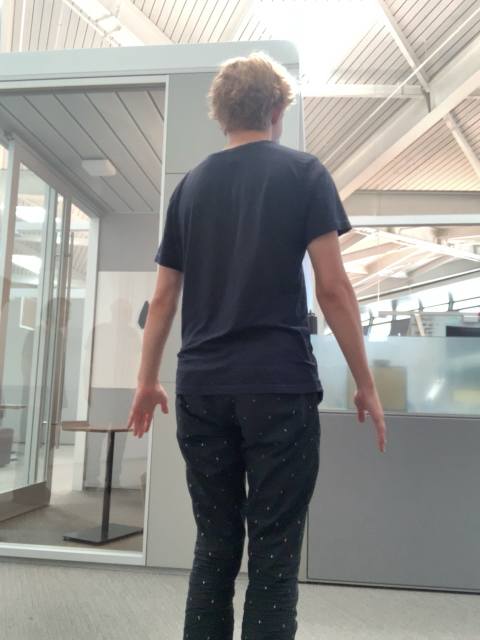} &
    \includegraphics[width=\mddaa]{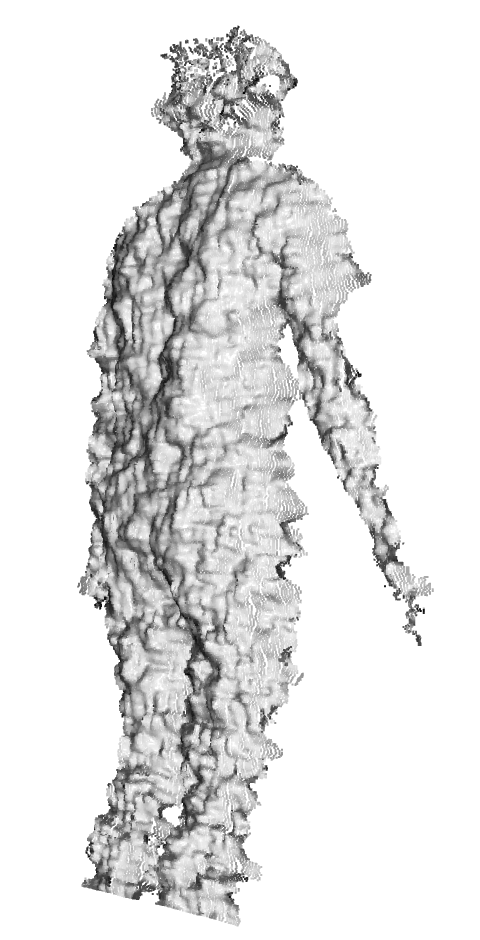} & 
    \includegraphics[width=\mddaa]{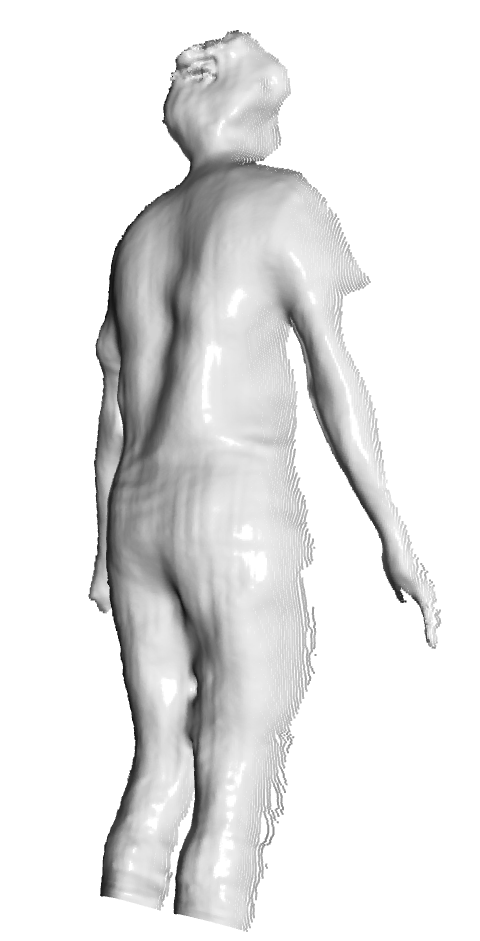} &
    \\ 
    
    \includegraphics[width=1.4cm]{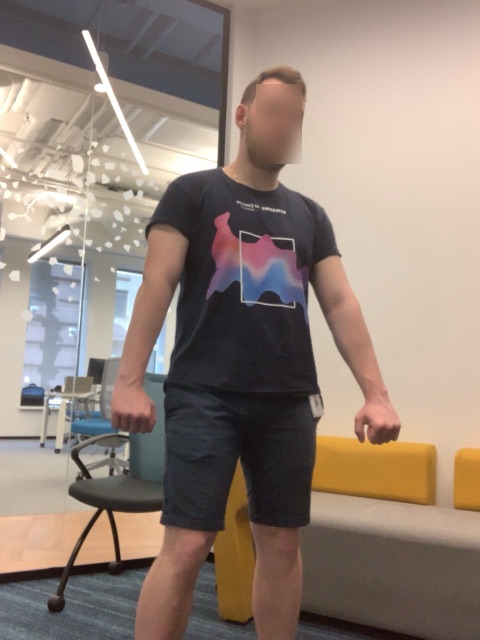} &
    \includegraphics[width=1.1cm]{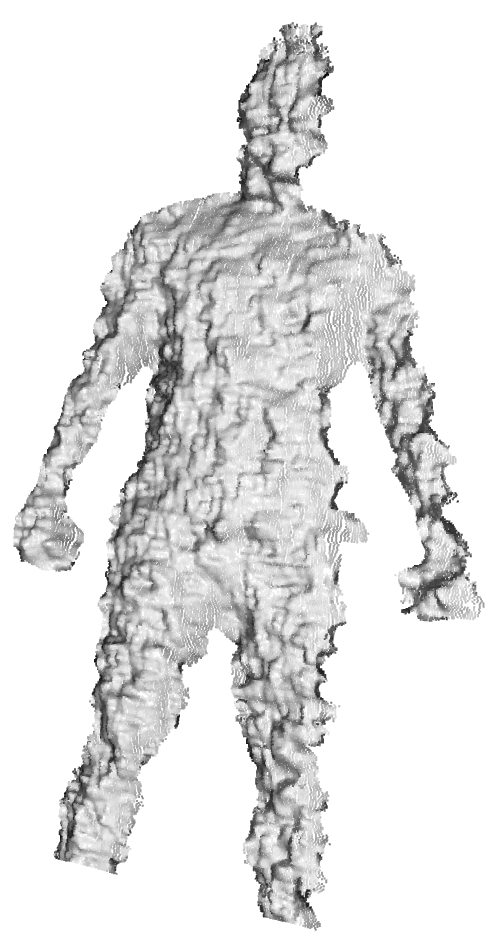} & 
    \includegraphics[width=1.1cm]{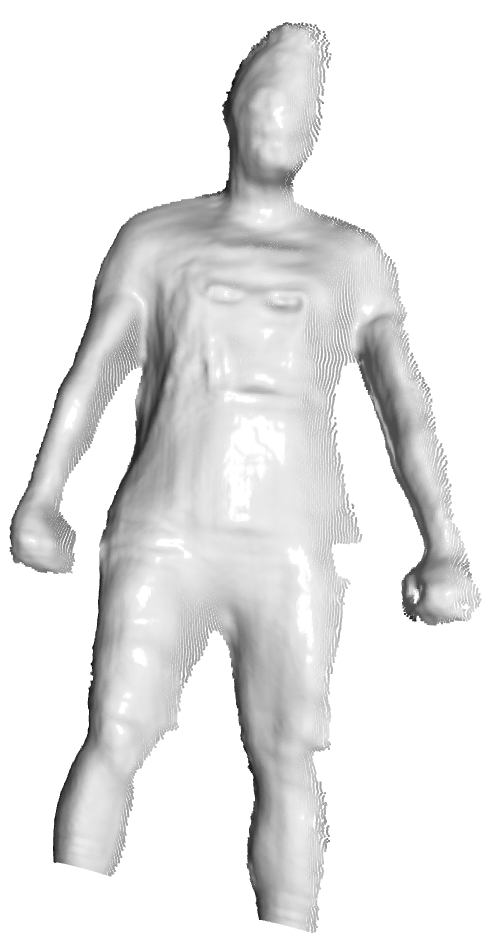} &
    
    \includegraphics[width=1.4cm]{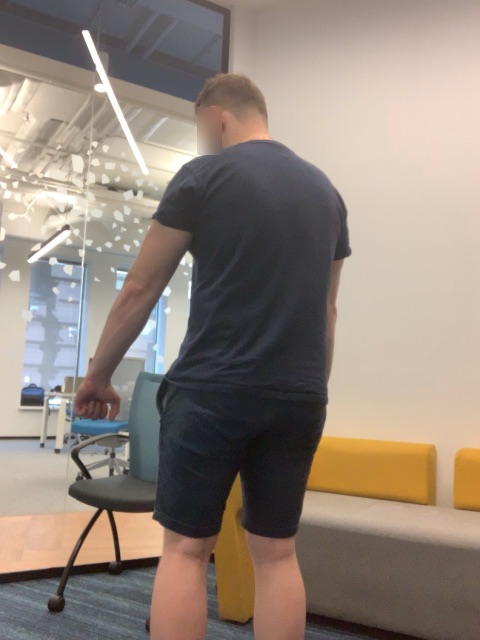} &
    \includegraphics[width=1.1cm]{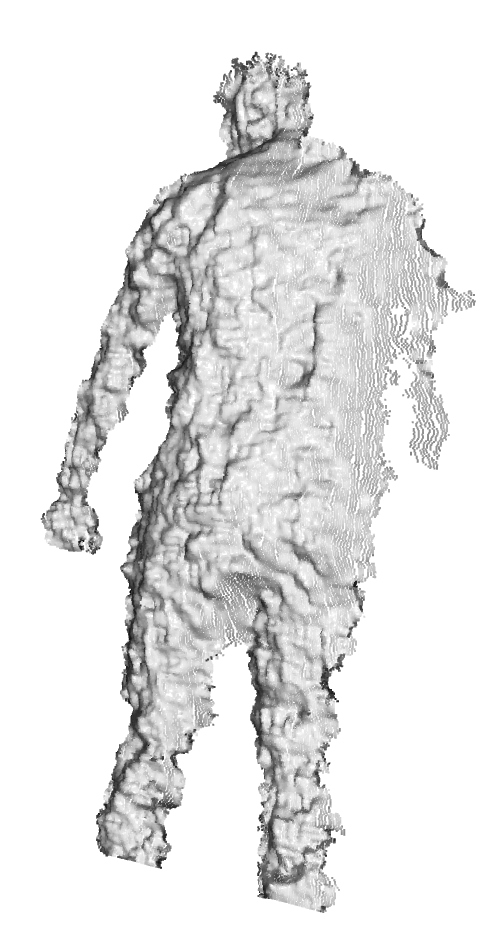} & 
    \includegraphics[width=1.1cm]{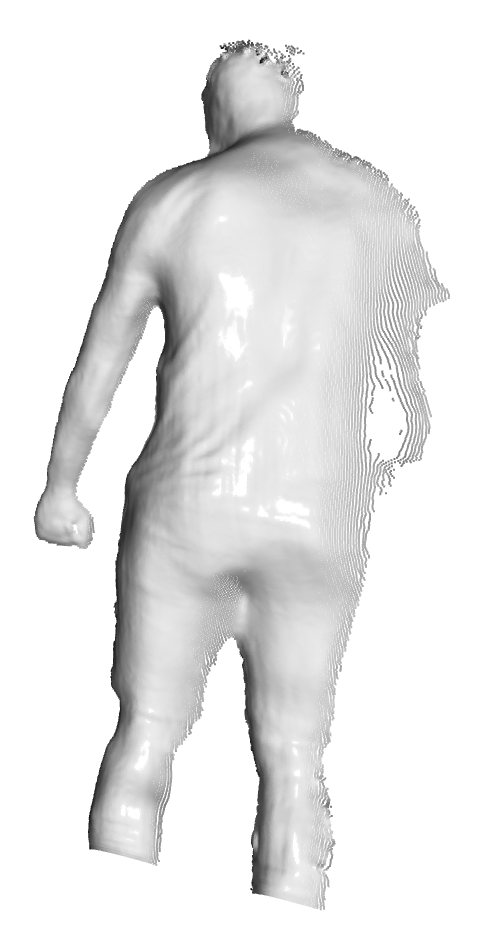} &
    \\ 
    \includegraphics[width=1.5cm]{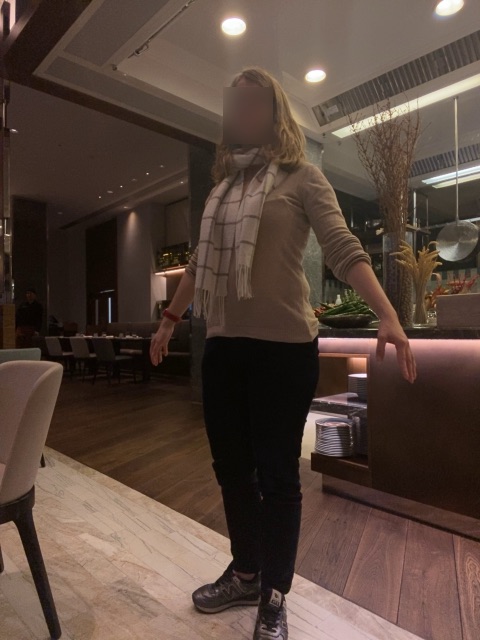} &
    \includegraphics[width=0.8cm]{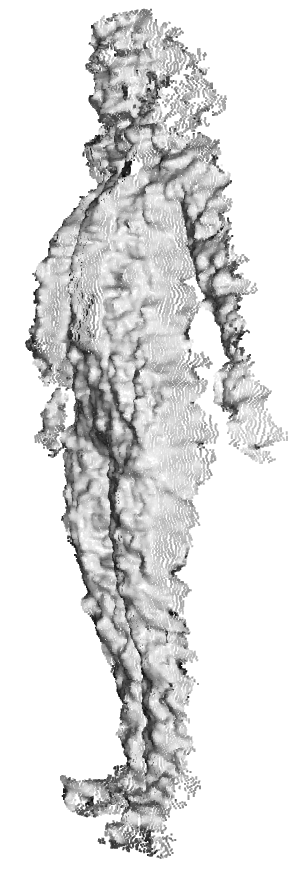} & 
    \includegraphics[width=0.8cm]{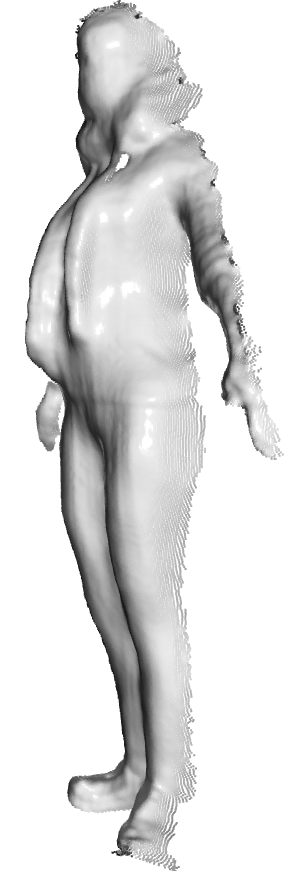} &
    
    \includegraphics[width=1.5cm]{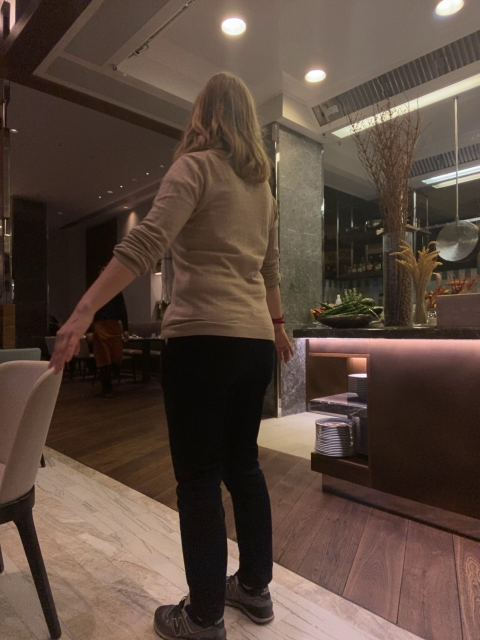} &
    \includegraphics[width=1.35cm]{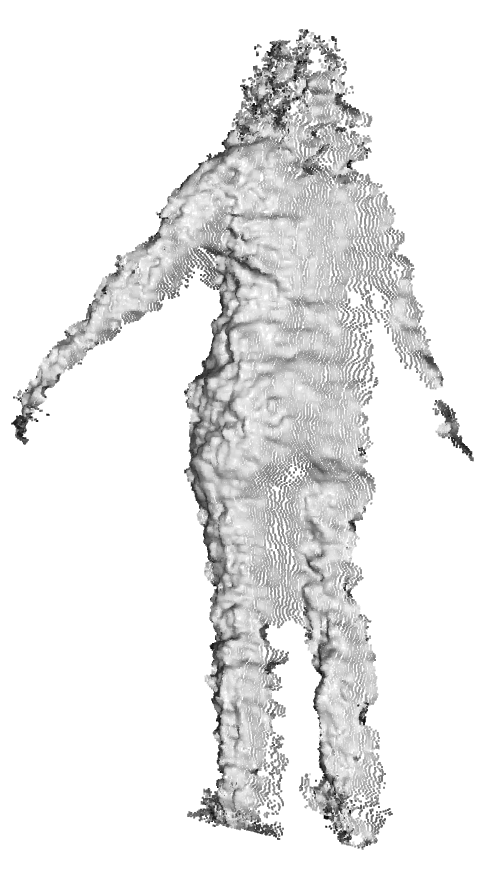} & 
    \includegraphics[width=1.3cm]{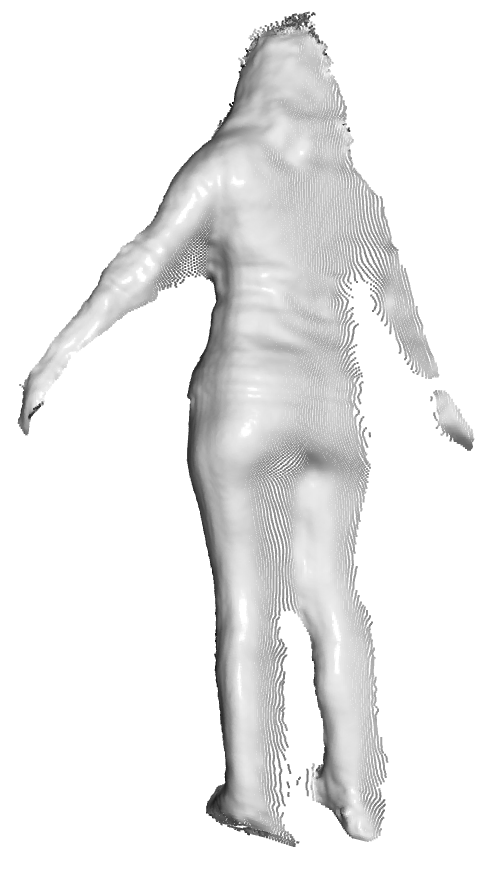} &
    
    \\
    \small{(a) Color} & \small{(b) TD} & \small{(c) Ours (basic)} & \small{(d) Color} & \small{(e) TD} & \small{(f) Ours (basic)}
    \end{tabular}}
    \vspace*{-1mm}\caption{More examples produced by our model for various people and environments.}\label{fig:comparison_lq_sumpat_2}
\end{figure*}

\begin{figure*}
    \centering
    \newlength{\mddsup}
    \setlength{\mddsup}{10mm}
    \centering\resizebox{1\linewidth}{!}{
    \begin{tabular}{@{}cccccccccc@{}}
    
    \includegraphics[width=12mm]{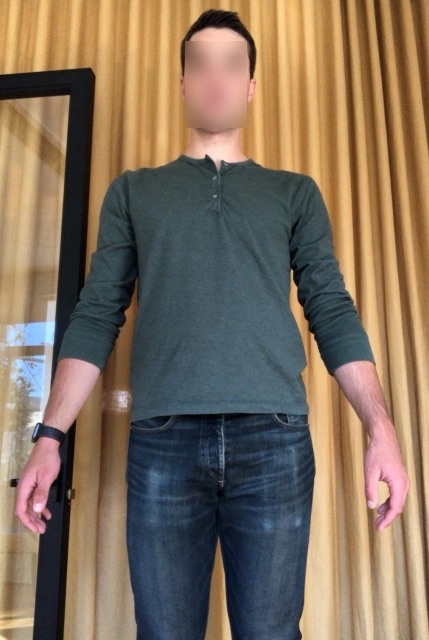}&
    \includegraphics[width=11mm]{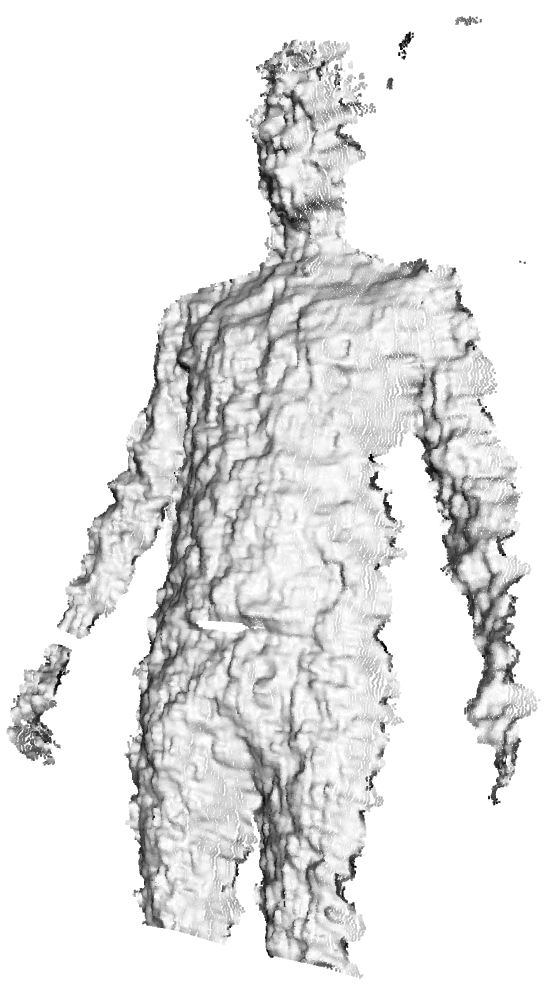} & 
    \includegraphics[width=11mm]{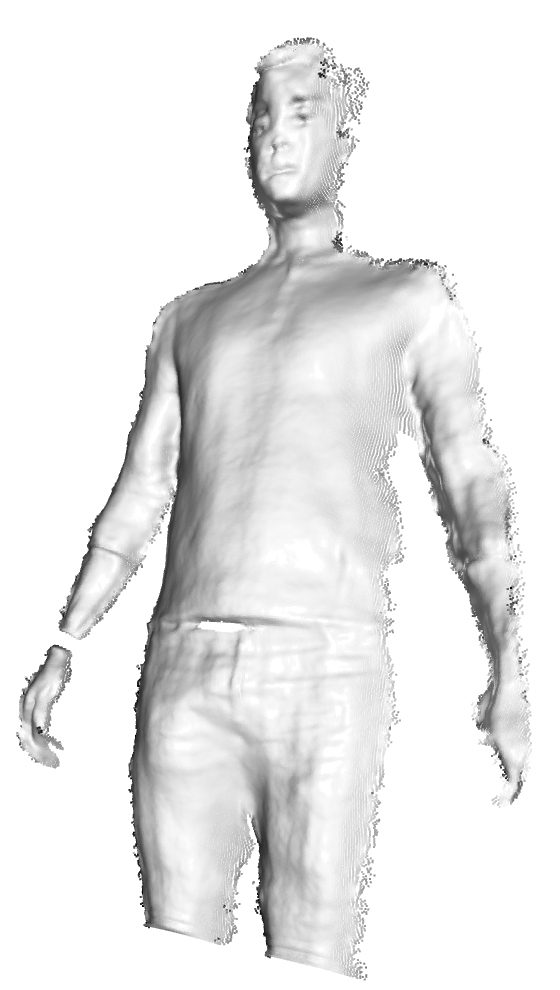} &
    \includegraphics[width=11mm]{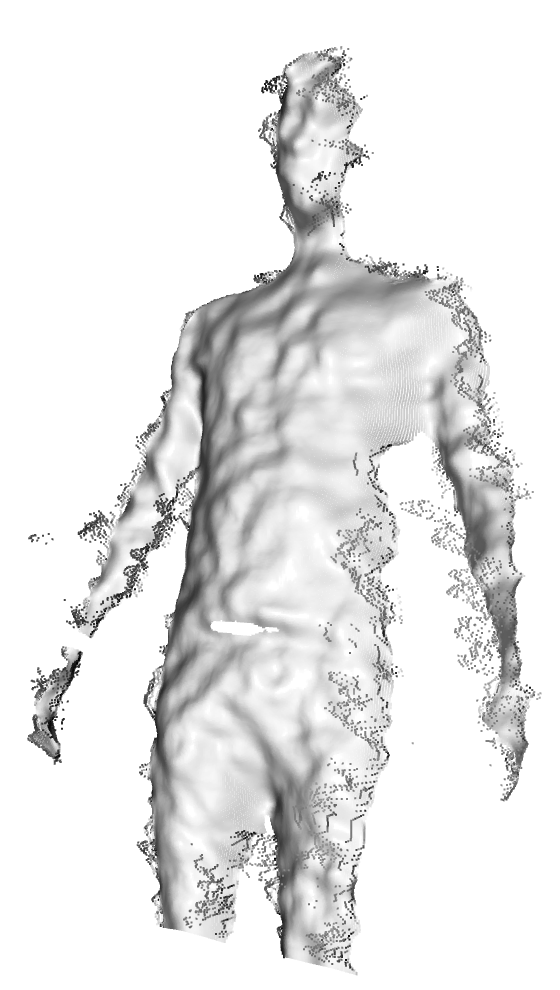}&
    \includegraphics[width=11mm]{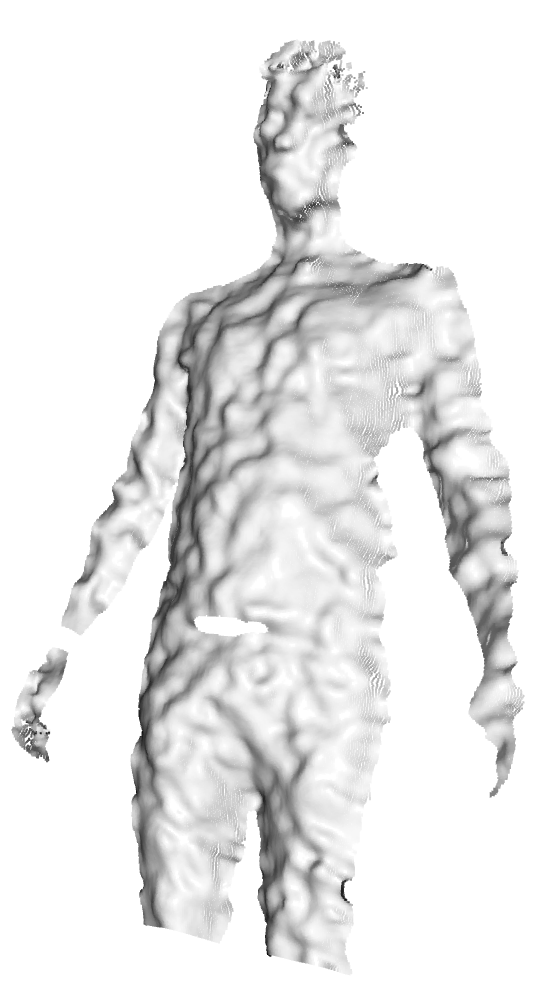} &
    \\ 
    
    \includegraphics[width=\mddsup]{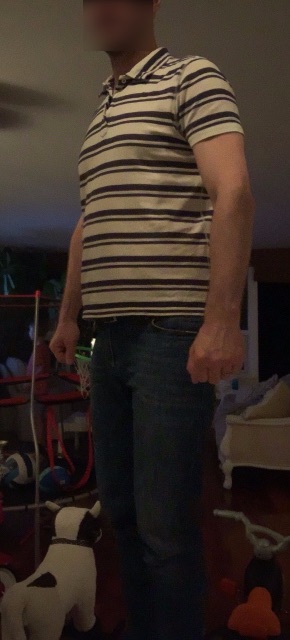}&
    \includegraphics[width=6mm]{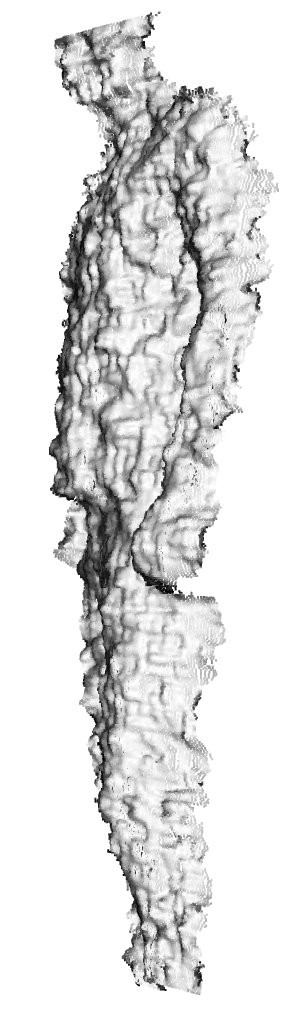} & 
    \includegraphics[width=6mm]{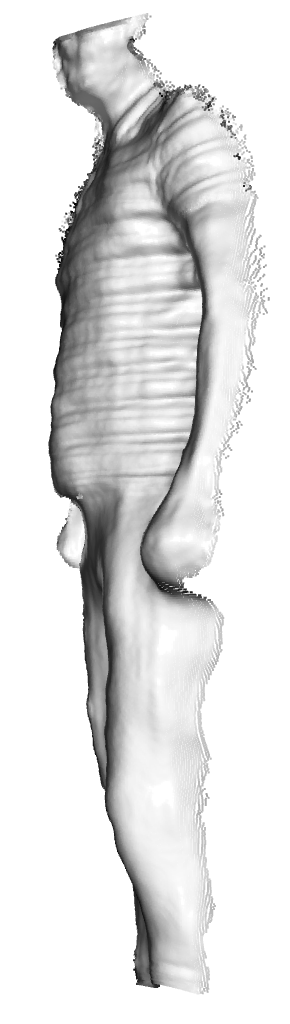} &
    \includegraphics[width=6mm]{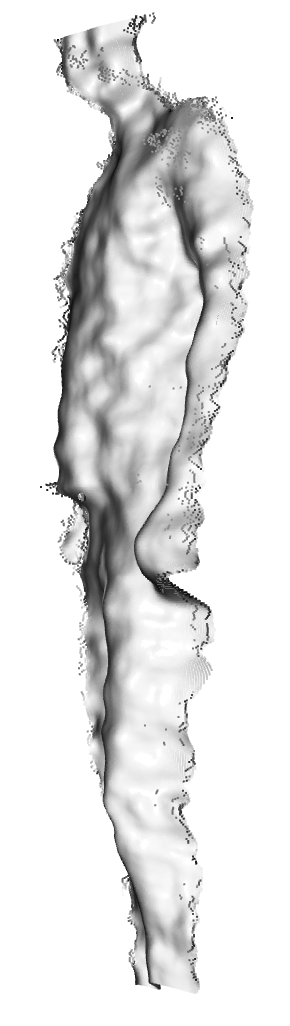}&
    \includegraphics[width=6mm]{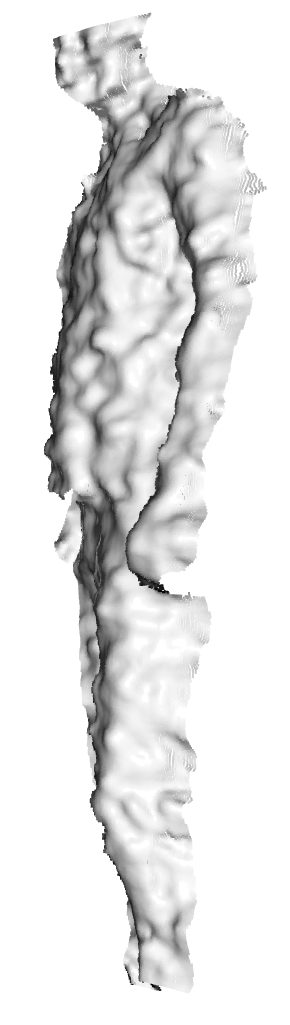} &
    \\ 
    
    \includegraphics[width=\mddsup]{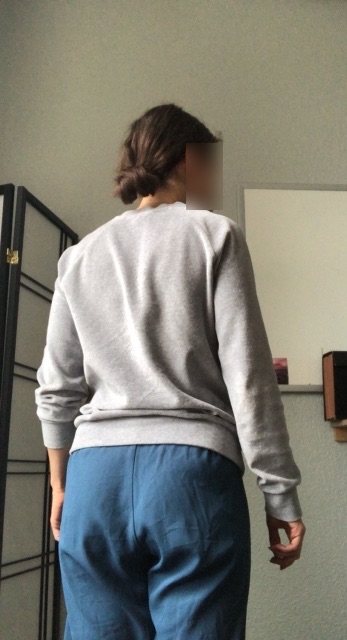}&
    \includegraphics[width=9mm]{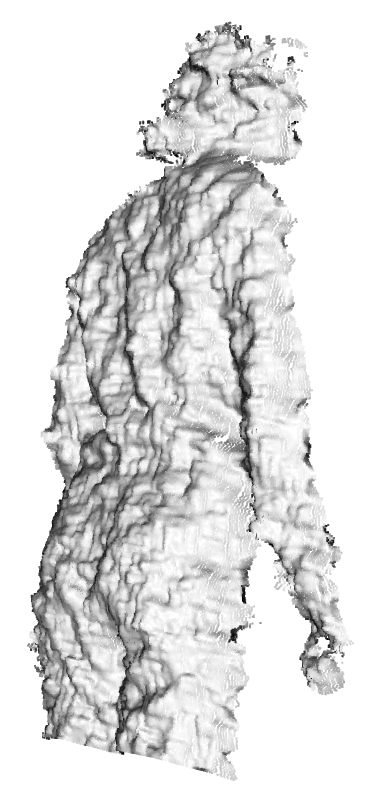} & 
    \includegraphics[width=9mm]{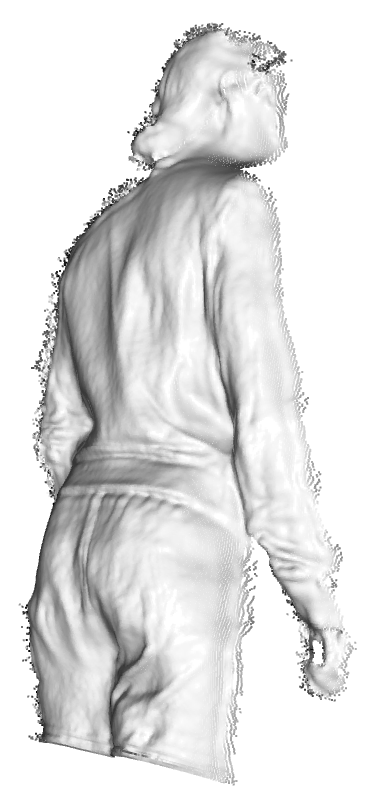} &
    \includegraphics[width=9mm]{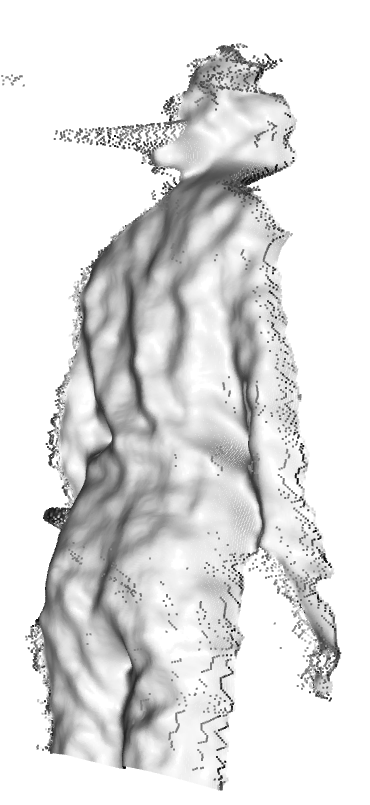}&
    \includegraphics[width=9mm]{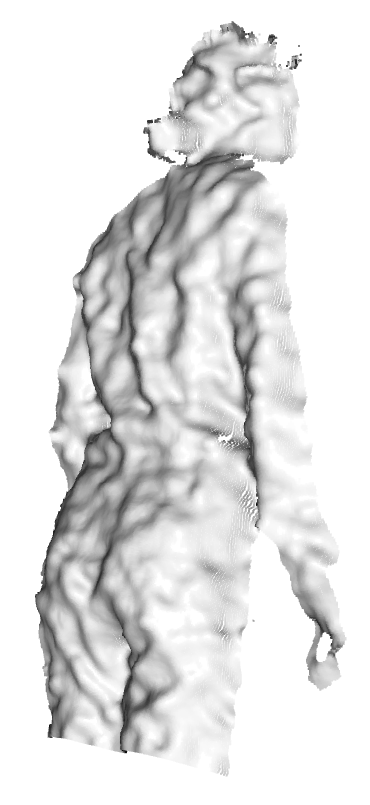} &
    \\ 
    
    \includegraphics[width=\mddsup]{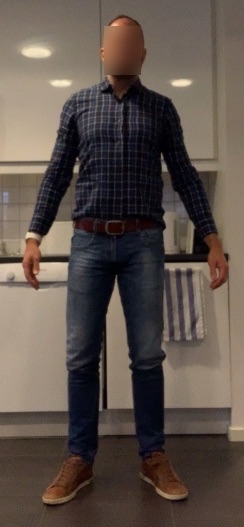}&
    \includegraphics[width=10mm]{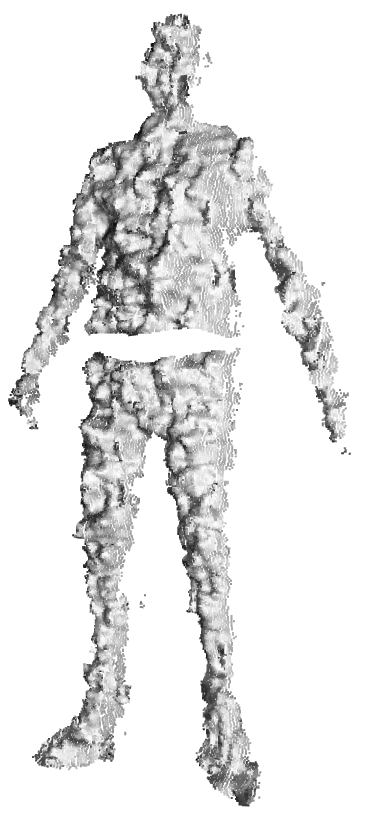} & 
    \includegraphics[width=10mm]{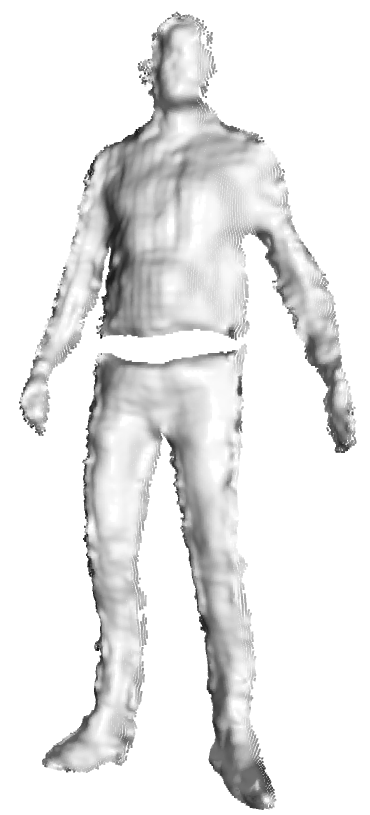} &
    \includegraphics[width=10mm]{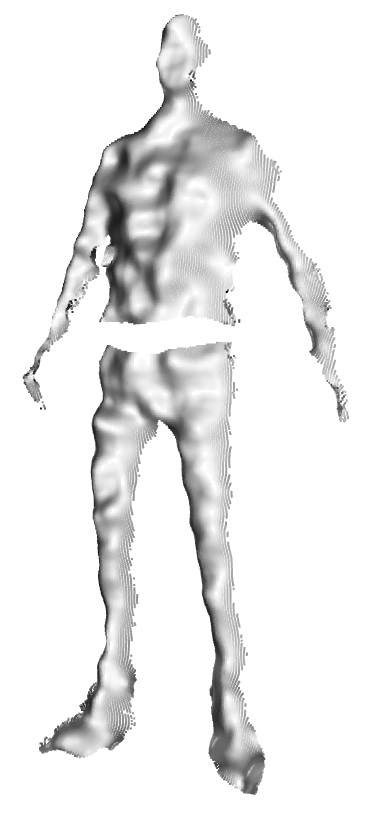}&
    \includegraphics[width=10mm]{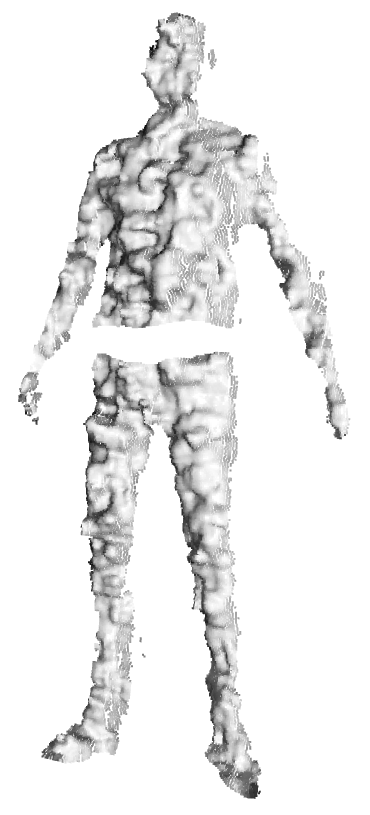} &
    \\ 
    
    \tiny{(a) Color} & \tiny{(b) TD} & \tiny{(c) Ours (basic)} & \tiny{(d) RGF} & \tiny{(e) DDRNet} 
    \end{tabular}}
    \vspace*{3mm}\caption{Qualitative comparison of depth denoising methods for lower-quality TD data. Here, we use our basic model as the input data is represented by separate RGB-D frames rather than sequences, and our full model cannot be applied.  }\label{fig:comparison_lq_sumpat}
\end{figure*}

In this supplementary material we provide more examples for qualitative assessment of our method~\cref{fig:supmat_big_compar,fig:comparison_lq_sumpat,fig:comparison_lq_sumpat_2}~. 

Here are \textbf{video} links with extensive comparison of our method. There are 2 video recordings:
\begin{itemize}
    \item \href{https://www.youtube.com/embed/29-tKQzmKJM}{Comparison of our methods}  - visualization of two proposed depth enhancement methods along with Kinect v2 and TrueDepth cameras recordings 
    \item \href{https://www.youtube.com/embed/cv8RpeSzVig}{Comparison of ours vs other methods}  - comparison of our basic method with filtering-based (RGF \cite{Jeon2018}) and data-driven (DRR \cite{Zhang2014rgf}) approaches.
\end{itemize}

\clearpage
{\small
\bibliographystyle{ieee}
\bibliography{egbib}
}

\end{document}